\documentclass[3p]{elsarticle}

\usepackage{graphicx}
\usepackage{fixmath}
\usepackage{adjustbox}
\usepackage[dvipsnames]{xcolor}
\usepackage{amssymb}
\usepackage{mathtools}
\usepackage[utf8]{inputenc} 
\usepackage[T1]{fontenc}    
\usepackage{hyperref}       
\usepackage{url}            
\usepackage{booktabs}       
\usepackage{amsfonts}       
\usepackage{nicefrac}       
\usepackage{subfloat}
\usepackage{lipsum}
\usepackage{subfig}
\usepackage{microtype}      
\usepackage{amsmath}
\usepackage{rotating}
\usepackage{array,multirow}
\usepackage{float}
\usepackage{lineno,hyperref}
\usepackage{natbib}
\usepackage[symbol]{footmisc}

\modulolinenumbers[5]

\makeatletter
\def\ps@pprintTitle{%
 \let\@oddhead\@empty
 \let\@evenhead\@empty
 \def\@oddfoot{}%
 \let\@evenfoot\@oddfoot}

\biboptions{authoryear}
\begin{document}

\begin{frontmatter}


\title{MILD-Net: Minimal Information Loss Dilated Network for Gland Instance Segmentation in Colon Histology Images\footnote[1]{This paper has been published at Medical Image Analysis. DOI: 10.1016/j.media.2018.11.009}}



\author{Simon Graham\textsuperscript{1,2}, Hao Chen\textsuperscript{3}, Jevgenij Gamper\textsuperscript{1,2}, Qi Dou\textsuperscript{3}, Pheng-Ann Heng\textsuperscript{3}, David Snead\textsuperscript{4}, Yee Wah Tsang \textsuperscript{4}, Nasir Rajpoot\textsuperscript{2,4,5}}

\address{\textsuperscript{1}Mathematics for Real World Systems Centre for Doctoral Training, University of Warwick, UK \\
\textsuperscript{2}Department of Computer Science, University of Warwick, UK \\
\textsuperscript{3}Department of Computer Science and Engineering, The Chinese University of Hong Kong, China \\
\textsuperscript{4}Department of Pathology, University Hospitals Coventry and Warwickshire, Coventry, UK \\
\textsuperscript{5}The Alan Turing Institute, London, UK}

\begin{abstract}

The analysis of glandular morphology within colon histopathology images is an important step in determining the grade of colon cancer. Despite the importance of this task, manual segmentation is laborious, time-consuming and can suffer from subjectivity among pathologists. The rise of computational pathology has led to the development of automated methods for gland segmentation that aim to overcome the challenges of manual segmentation. However, this task is non-trivial due to the large variability in glandular appearance and the difficulty in differentiating between certain glandular and non-glandular histological structures. Furthermore, a measure of uncertainty is essential for diagnostic decision making. To address these challenges, we propose a fully convolutional neural network that counters the loss of information caused by max-pooling by re-introducing the original image at multiple points within the network. We also use \textit{atrous} spatial pyramid pooling with varying dilation rates for preserving the resolution and multi-level aggregation. To incorporate uncertainty, we introduce random transformations during test time for an enhanced segmentation result that simultaneously generates an uncertainty map, highlighting areas of ambiguity. We show that this map can be used to define a metric for disregarding predictions with high uncertainty. The proposed network achieves state-of-the-art performance on the GlaS challenge dataset and on a second independent colorectal adenocarcinoma dataset. In addition, we perform gland instance segmentation on whole-slide images from two further datasets to highlight the generalisability of our method. As an extension, we introduce MILD-Net$^+$ for simultaneous gland and lumen segmentation, to increase the diagnostic power of the network.
\end{abstract}


\end{frontmatter}


\section{Introduction}
Colorectal cancer is the third most commonly occurring cancer in men and the second most commonly occurring cancer in women, where approximately 95\% of all colorectal cancers are adenocarcinomas~\citep{fleming2012colorectal}. Colorectal adenocarcinoma develops in the lining of the colon or rectum, which makes up the large intestine and is characterised by glandular formation. Histological examination of the glands, most frequently with the Hematoxylin \& Eosin (H\&E) stain, is routine practice for assessing the differentiation of the cancer within colorectal adenocarcinoma. Pathologists use the degree of glandular formation as an important factor in deciding the grade or degree of differentiation of the tumour. Within well differentiated cases, above 95\% of the tumour is gland forming~\citep{fleming2012colorectal}, whereas in poorly differentiated cases, typical glandular appearance is lost. Within the top row of Figure 1, (a) shows a healthy case, (b) shows a moderately differentiated tumour and (c) shows a poorly differentiated tumour. We observe the loss of glandular formation as the grade of cancer increases. 


\begin{figure}[!t]
\captionsetup[subfigure]{labelformat=empty}
\centering
\subfloat[]{\includegraphics[width=0.95\columnwidth]{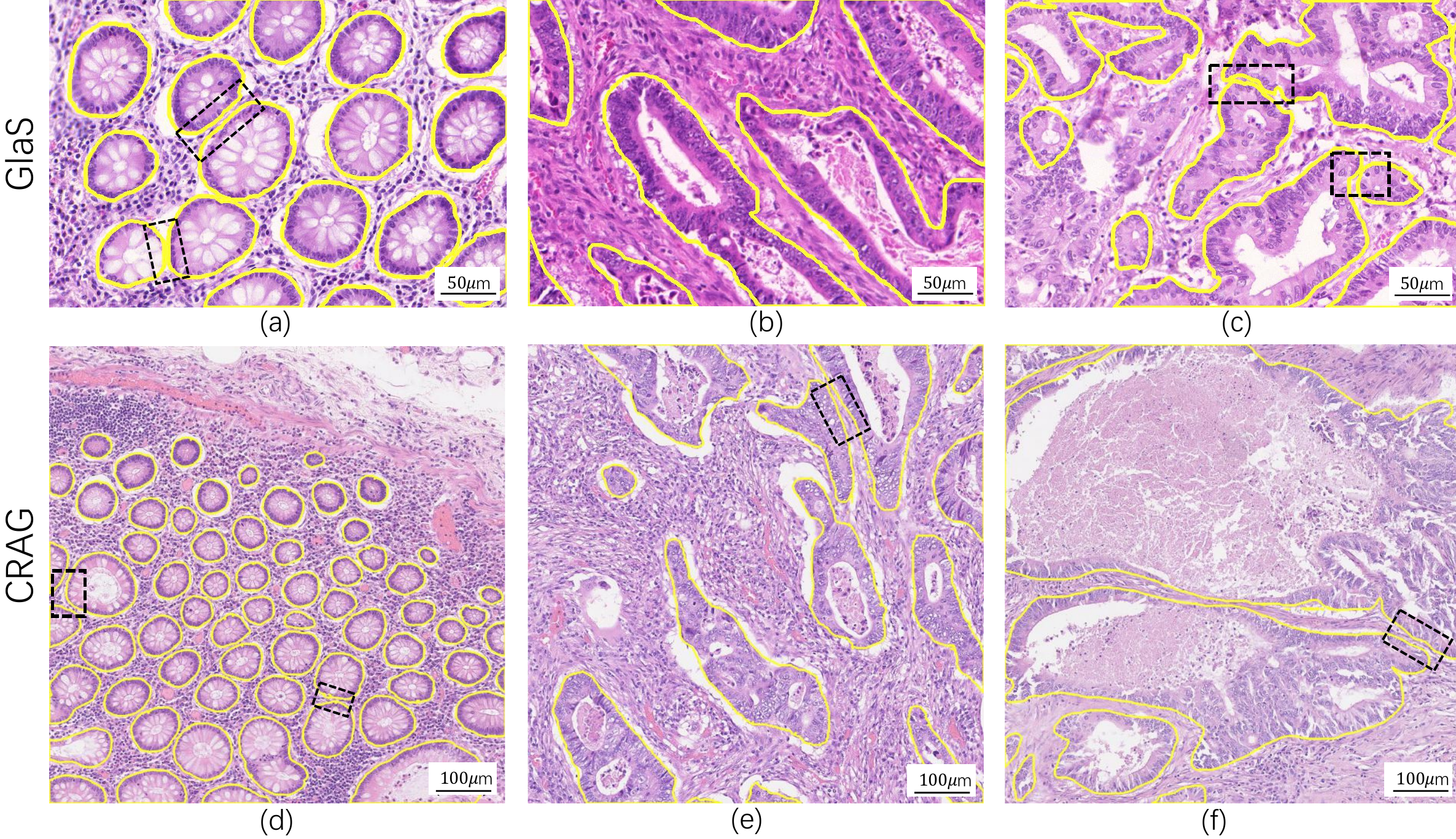} \label{fig:detection1}}
\caption{(a-c) Example images from the GlaS dataset~\citep{sirinukunwattana2017gland}. (d-f) Example images from the CRAG dataset. All images displayed have overlaid boundary ground truth as annotated by an expert pathologist and are at 20$\times$ magnification. (a) and (d) show healthy glands, whereas the other images contain malignant glands. Black boxes highlight clustered glands.
}
\label{fig:detection}
\end{figure}

There is a growing trend towards a digitised pathology workflow, where digital images are acquired from glass histology slides using a scanning device. The advent of digital pathology has led to a rise in computational pathology, where algorithms are implemented to assist pathologists in diagnostic decision making. In routine pathological practice, accurate segmentation of structures such as glands and nuclei are of crucial importance because their morphological properties can assist a pathologist in assessing the degree of malignancy~\citep{compton2000updated,hamilton2000pathology,washington2009protocol}. With the advent of computational pathology, digitised histology slides are being leveraged such that pathological segmentation tasks can be completed in an objective manner. In particular, automated gland segmentation within H\&E images can enable pathologists to extract vital morphological features from large scale histopathology images, that would otherwise be impractical.

Computerized techniques play a significant role in automated digitalized histology image analysis, with applications to various tasks including but limited to nuclei detection and segmentation~\citep{graham2018sams,chen2017dcan,sirinukunwattana2016locality}, mitosis detection~\citep{cirecsan2013mitosis,chen2016mitosis,veta2015assessment,albarqouni2016aggnet}, tumor segmentation~\citep{qaiser2017tumor}, image retrieval~\citep{sapkota2018deep,shi2017supervised}, cancer type classification~\citep{graham2018classification,kong2017cancer,bejnordi2017diagnostic,lin2018scannet,qaiser2018her}, etc.
Most of the previous literature focused on the hand-crated features for histopathological image analysis~\citep{gurcan2009histopathological}. 
Recently, deep learning achieved great success on image recognition tasks with powerful feature representation~\citep{litjens2017survey,shen2017deep,lecun2015deep}. 
For example, U-Net achieved excellent performance on the gland segmentation task~\citep{ronneberger2015u}. To further improve the gland instance segmentation performance, Chen et al. presented a deep contour-aware network by formulating an explicit contour loss function in the training process and achieved the best performance during the 2015 MICCAI Gland Segmentation (GlaS) on-site challenge~\citep{chen2016dcan,chen2017dcan,sirinukunwattana2017gland}.
In addition, a framework was proposed in~\cite{xu2016gland} by fusing complex multichannel regional and boundary patterns with side supervision for gland instance segmentation. This work was extended in~\cite{xu2017gland} to incorporate additional bounding box information for an enhanced performance.
A Multi-Input-Multi-Output network (MIMO-Net) was presented for gland segmentation in~\cite{raza2017mimonet} and achieved the state-of-the-art performance.
Furthermore, several methods have investigated the segmentation of glands from histology images using limited expert annotation effort. For example, a deep active learning framework was presented in~\cite{yang2017suggestive} for gland segmentation using suggestive annotation. Unannotated images were utilized in~\cite{zhang2017deep} with the design of deep adversarial networks and consistently good segmentation performance was attained.

However, automated gland segmentation remains a challenging task due to several important factors. First, a high resolution level is needed for precise delineation of glandular boundaries, that is important when extracting morphological measurements.  Next, glands vary in their size and shape, especially as the grade of cancer increases. Furthermore, the output of solely the gland object gives limited information when making a diagnosis. Extra information, such as the uncertainty of a prediction and the simultaneous segmentation of additional histological components, may give additional diagnostic power. For example, the pathologist may choose to ignore areas with high uncertainty, such as areas with dense nuclei and areas containing artifacts. An additional histological component of particular interest is the lumen, which is ultimately the defining structure of a gland. This structure can help empower diagnostic decision making, because its presence and morphology can be indicative of the grade of cancer.

In this paper we propose a minimal information loss dilated network that aims to solve the key challenges posed by automated gland segmentation. During feature extraction, we introduce minimal information loss (MIL) units, where we incorporate the original down-sampled image into the residual unit after max-pooling. This, alongside dilated convolution, helps retain maximal information that is essential for segmentation, particularly at the glandular boundaries. We use \textit{atrous} spatial pyramid pooling for multi-level aggregation that is essential when segmenting glands of varying shapes and sizes. After feature extraction, our network up-samples the feature maps to localise the regions of interest. During uncertainty quantification, we apply random transformations to the input images as a method of generating the predictive distribution. This leads to a superior segmentation result and allows us to observe areas of uncertainty that may be clinically informative. Furthermore, we use this measure of uncertainty to rank images that should be prioritised for pathologist annotation. As an extension, we demonstrate how our method can be modified to simulatenously segment the gland lumen. The additional segmentation of the gland lumen can empower current automated methods to achieve a more accurate diagnosis. 

Our proposed framework can be trained end to end, with one minimal information loss dilated feature extraction network. Experimental results show that the proposed framework achieves state-of-the-art performance on the 2015 MICCAI GlaS Challenge dataset and on a second independent colorectal adenocarcinoma dataset.

\begin{figure}[!t]
\centering
\captionsetup[subfigure]{labelformat=empty}
\subfloat[]{\includegraphics[width=0.99\columnwidth]{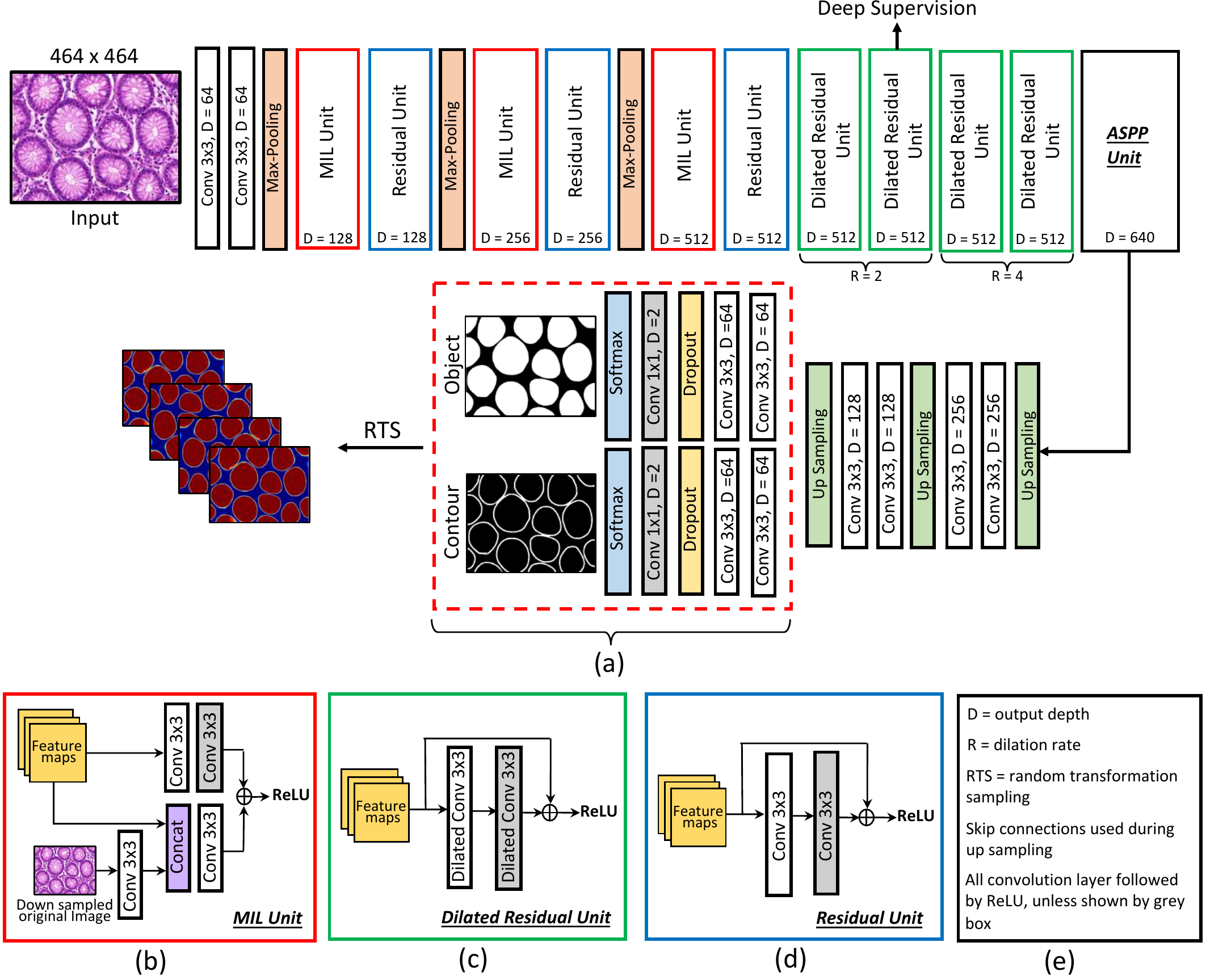} \label{fig:framework}}
\caption*{Figure 2: The overall framework of the proposed method. (a) Task specific component of the network. We show in section 2.3 how this component can be modified. (b-d) lllustration of the varying residual units. (e) Key showing important components of the framework.}
\label{fig:detection}
\end{figure}

\section{Methods}
\subsection{Minimal Information Loss Dilated Network}
Gland instance segmentation is a complex task that requires a significantly deep network for meaningful feature extraction. Therefore, we use residual units to allow efficient gradient propagation through our deep network architecture. Traditional convolutional neural networks use a combination of max-pooling and convolution in a hierarchical fashion to increase the size of the receptive field~\citep{lecun2015deep}. The inclusion of max-pooling results in the loss of information with relatively low activations \citep{sabour2017dynamic}, that is important for pixel-level prediction in segmentation. A significant amount of downsampling via max-pooling leads to a sub-optimal segmentation, particularly at thin object boundaries and for small objects. To counter this loss of information, in addition to using traditional residual units,  we include two additional types of residual units during feature extraction: MIL units and dilated residual units. The MIL unit incorporates the original image into each residual unit directly after the max-pooling layer. First, the original image is down-sampled to the same size as the output of the pooling operation by bicubic interpolation. Then, a 3$\times$3 convolution is applied before concatenating to the output of the pooling layer. Next, a 3$\times$3 convolution is applied to the concatenated block and this output is subsequently used in the residual summation operation, as opposed to the input tensor in traditional methods. Three MIL units are added during feature extraction immediately after max-pooling. These MIL units can be seen in more detail within part (a) of Figure 2. A traditional residual unit, which is defined as:

\begin{equation}
\textbf{y} = \mathcal{F}(\textbf{x} , {\textbf{W}}) + \textbf{x}
\end{equation}

\noindent where $\textbf{x}$ and $\textbf{y}$ denote the input and output vectors respectively and $\textbf{W}$ denotes the weights within the residual unit. Specifically $\mathcal{F}$ represents the function $\textbf{W}_2(\sigma(\textbf{W}_1\textbf{x}))$, where $\sigma$ denotes ReLU, $\textbf{W}_1$ denotes the weights of the first convolution and $\textbf{W}_2$ denotes the weights of the second convolution. The addition of the the input vector \textbf{x} to $\mathcal{F}$ is shown by the summation operator $\oplus$ in the residual unit of part (d) in Figure 2. When we use a downsampled version of the original image (downsampled with bicubic interpolation) without max-pooling, it indirectly captures the variation in pixel intensities in the local neighbourhood of each pixel without completely discarding the activations, as is the case with max-pooling. It is this principle that allows the MIL unit to ensure that the missing details are preserved. Equation (1) is modified to generate the MIL unit. The MIL unit can be defined as:

\begin{equation}
\textbf{y} = \mathcal{F}(\textbf{x} , {\textbf{W}}) + \mathcal{G}(\textbf{x} , \textbf{v}, {\textbf{M}})
\end{equation}

\noindent where $\mathcal{F}$ is defined in the same way as equation (1). The vector $\textbf{v}$ denotes the original down-sampled image and is incorporated into the function $\mathcal{G}$ to minimise the loss of information. $\mathcal{G}$ represents the function  $\textbf{M}_2(\sigma(\textbf{M}_1\textbf{v}) \Vert \textbf{x})$, where $\Vert$ denotes the concatenation operation. Similar to the the traditional residual unit,  $\textbf{M}_1$ and $\textbf{M}_2$ within function $\mathcal{G}$ represent the weights of the convolution applied to the down-sampled image and the convolution of the concatenated feature maps respectively. The summation of $\mathcal{F}$ and $\mathcal{G}$ is shown by the  $\oplus$ symbol in the MIL unit within Figure 2.

Instead of downsampling the size of the input to increase the size of the receptive field, an alternate solution is to increase the size of the kernel during convolution. However, this practice is not feasible due to the huge amount of parameters required. Instead, dilated convolution uses sparse kernels~\citep{yu2015multi}, such that the resolution of the original image is preserved, without significantly increasing the number of parameters. We incorporate dilated convolution into residual units simply by replacing each 3$\times$3 convolution with a 3$\times$3 dilated convolution. We initially down-sample using max-pooling and MIL units and then use dilated convolution when the image has been down-sampled by a factor of 8. We do not use dilated convolution throughout the entire network since otherwise the model does not fit into GPU memory. This is because convolving over the size of the original image requires a greater amount of parameters compared to when this image is down-sampled. Dilated residual units can be seen in part (b) of Figure 2. Minimising the loss of information allows us to perform a successful gland instance segmentation, without the need to incorporate additional information that is used in other methods~\citep{chen2017dcan}. Retaining the information throughout the model allows the network to successfully segment small glandular objects and thin glandular contours. It must be noted that we output the contours for uncertainty map refinement; not for separating gland instances. This is explained further in section 2.2. 

In addition, for effective multi-level aggregation, we apply \textit{atrous} spatial pyramid pooling (ASPP)~\citep{chen2018deeplab} to the output of the deep network. Within our framework, the goal of ASPP is to combat the challenge of detecting glands of different cancer grades that display a high level of morphological heterogeneity. To achieve this, we merge together multiple dilated convolution layers, allowing us to explicitly control the size of the receptive field. Specifically, we use three dilated convolution operations, with rates 6, 12 and 18. When the dilation rate is large, the dilated convolution reduces to a 1$\times$1 convolution. This is because the dilated kernel becomes larger than the size of the input feature map. Instead, to incorporate global level context, we also use global average pooling. All operations are followed by an initial 1$\times$1 convolution, a dropout layer with a rate of 0.5 and then a second 1$\times$1 convolution for reducing the depth of the output. The concatenation of these feature maps gives a powerful representation of the features extracted from the minimal information loss dilated network.

Although high-level contextual information can be generated within the deep neural network, it is crucial to incorporate low-level information for precisely delineating the glandular boundaries. Directly upsampling by a factor of 8 to produce the output does not consider low-level information. Instead, similar to U-Net~\citep{ronneberger2015u}, we choose to up-sample by a factor of 2 each time and concatenate low-level features to the start of each upsampling block. Before the concatenation, we apply a 1$\times$1 convolution to increase the depth of lower levels; ensuring that we have an equal contribution of both components during the concatenation. We concatenate the feature maps from the second convolution layer and the first two standard residual units. We find that this method of upsampling is especially important for precisely locating the boundaries where low-level features are particularly important. When the features have been up-sampled to the resolution of the original image, the network splits into two separate branches: one for the gland object and one for the contour. We denote this part of the network the task specific component of the network and is shown by the dashed red box in Figure 2(a). We show an example of how the task specific component can be modified in section 2.3. We add deep supervision to our network by calculating the auxiliary loss at the second dilated residual unit during feature extraction. This helps the network to learn more discriminative features and encourages a faster convergence. We also add dropout layers immediately before the final 1$\times$1 convolution, near the output of the network, with a rate of 0.5. The overall flow of the network can be seen in Figure 2. 

\subsection{MILD-Net Loss Function}

During training, we calculate the cross-entropy loss with respect to all outputs of the proposed network. Concretely, we define $\mathcal{L}_g$, $\mathcal{L}_c$, $\mathcal{L}_{a_g}$, $\mathcal{L}_{a_c}$ to be the gland, contour, gland auxiliary and contour auxiliary cross-entropy loss respectively and are formally given below in equation (3).

\begin{equation}
\begin{split} \\
&\mathcal{L}_g = -\sum_{x \in \chi}{\log p_g(x, y_g(x);  \mathbold{w}_g)}\\ 
&\mathcal{L}_c = -\sum_{x \in \chi}{\log p_c(x, y_c(x);\mathbold{w}_c)}\\
&\mathcal{L}_{a_g} = -\sum_{x \in \chi}{\log p_{a_g}(x, y_{a_g}(x);\mathbold{w}_{a_g})}\\
&\mathcal{L}_{a_c} = -\sum_{x \in \chi}{\log p_{a_c}(x, y_{a_c}(x);\mathbold{w}_{a_c})}
\end{split}
\end{equation}

Here, $p_g(x, y_g(x);\mathbold{w}_g)$, $p_c(x, y_c(x);\mathbold{w}_c)$,  $p_{a_g}(x, y_{a_g}(x);\mathbold{w}_{a_g})$ and  $p_{a_c}(x, y_{a_c}(x);\mathbold{w}_{a_c})$ represent the pixel-based softmax classification at the gland, contour, auxiliary gland and auxiliary contour output for true labels $y_g(x), y_c(x), y_{a_g}(x)$ and  $y_{a_c}(x)$ respectively. $x$ denotes a given input pixel in image space $\chi$. To obtain the overall loss for each component, the sum of the cross-entropy loss for each image is calculated. Then, the overall loss function to be minimised during training is defined as:

\begin{equation}
\mathcal{L}_{total} = \mathcal{L}_g + \mathcal{L}_c + \lambda\mathcal{L}_{a_g} + \lambda\mathcal{L}_{a_c} + \gamma||\mathbold{w}||_2^2
\end{equation}
\noindent where discount weight $\lambda$ decays the contribution of each auxiliary loss $\mathcal{L}_{a_g}$ and $\mathcal{L}_{a_c}$ during training.  We initially set $\lambda$ as 1, and divide the value by 10 after every eight training epochs. The selection of the initial $\lambda$ and the decay strategy was motivated by DCAN~\cite{chen2017dcan}, where they used a similar strategy. $||\mathbold{w}||_2^2$ denotes the regularisation term on weights $\mathbold{w}=\{\mathbold{w}_g,\mathbold{w}_c,\mathbold{w}_{a_g},\mathbold{w}_{a_c}\}$, with regularisation parameter $\gamma$. We emperically set gamma to be $10^{-5}$.

\subsection{Random Transformation Sampling for Uncertainty Quantification}

Current deep learning models have an ability to learn powerful feature representations and are capable of successfully mapping high dimensional input data to an output. However, this mapping is assumed to be accurate in such models and there is no quantification of how certain the model is of the prediction. Bayesian approaches to modeling, naturally involve uncertainty quantification by obtaining a posterior distribution over the parameters of the model, which therefore allows us to induce a predictive distribution for the unseen data. However, the tractability and scalability of Bayesian methods applied to shallow neural networks and their recent deeper counterparts have been a subject of research for the past several decades. Although significant progress has been made, inference of the posterior distribution over the model parameters remains computationally expensive. Recent work~\citep{gal2016dropout} demonstrated that using a standard regularisation tool such as dropout is equivalent to variational approximation using Bernoulli distributions~\citep{bishop2012pattern} in deep learning. Therefore, this can be used to approximate the uncertainty over the model predictions~\citep{Gal2016Uncertainty}. Standard variational dropout captures the uncertainty over the model weights, given the observed data. It is important to distinguish that there may be noise inherent to each observation, that we might not be able to reduce by obtaining more data. This would be crucial to estimate within clinical applications. Generally, this uncertainty is estimated through a data dependent noise model~\citep{kendall2017uncertainties}, however it would require us to modify the existing architecture. Therefore, to capture observation dependent noise, we perform random transformations to the input images during test time. To obtain the predictive distribution, we apply a random transformation $\Phi(\textbf{x})$ on a sample of $n$ images, where $\Phi$ performs a flip, rotation, Gaussian blur, median blur or adds Gaussian noise on input image $\textbf{x}$ to obtain $\{\Phi_1, \Phi_2, ... , \Phi_n\}$. Each image within the sample is then processed, where the mean of this processed sample gives the refined prediction and the variance gives the uncertainty. Due to the aggregation of the predictions of multiple transformed images, our model will naturally perform well, particularly for areas that are generally difficult to classify. Similarly, recent work leveraged transformed images, but instead are utilised to obtain informative priors \citep{nalisnick2018learning}, that help a model become more invariant to these specific transformations. However, the primary aim for utilising RTS is to obtain a measure of uncertainty that may be informative within clinical practice, as opposed to making our model more invariant. We can define the prediction and uncertainty as:

\begin{equation}
\mu= \frac{1}{n}\sum_{i=1}^{n}{f(\Phi_i(\textbf{x});\mathbold{w})};~~~\sigma = \frac{1}{n}\sum_{i=1}^{n}{(f(\Phi_i(\textbf{x});\mathbold{w})}-\mu)^2
\end{equation}

\noindent where $\mu$ defines the segmentation prediction, $\sigma$ defines the uncertainty and $n$ defines the number of transformations. The function $f$ denotes the deep neural network with input \textbf{x} and output taken after the softmax layer. $\mathbold{w}$ denotes the weights and $\Phi_i$ defines a random transformation $i$ to input image $\textbf{x}$. Note, that the output of $\sigma$ is a two-dimensional image, where high values denote pixels with high uncertainty. 

We propose a metric to give individual glands a score of uncertainty, based on the uncertainty map generated via random transformation sampling (RTS). This measure highlights glands that are generally hard to classify, irrespective of the number training examples that the model has seen. We suggest that it is reasonable to disregard segmented glands that have an uncertainty score above a given threshold, because in practice features would not be extracted from areas of general ambiguity. We first remove the boundaries by subtracting the predicted contours that have been output by the network and then calculate the object-level uncertainty score for each predicted instance $k$ as: $\tau_{k}$ = $\frac{1}{n}\sum_{i=1}^{n} \hat{\sigma}\rho_{k,i} $, where $\hat{\sigma}$ is the boundary removed uncertainty map and $\rho_{k,i}$ is the predicted binary output of pixel $i$ within instance $k$. We define $n$ as the number of pixels within predicted instance $k$. We remove the boundaries because these areas show the transition between the two classes and therefore the uncertainty here can't be avoided. Given a selected global threshold for our uncertainty score $\tau$, we may only consider segmented glands with a score below this threshold. 

\begin{figure}[!t]
\centering
\captionsetup[subfigure]{labelformat=empty}
\subfloat[]{\includegraphics[width=0.8\columnwidth]{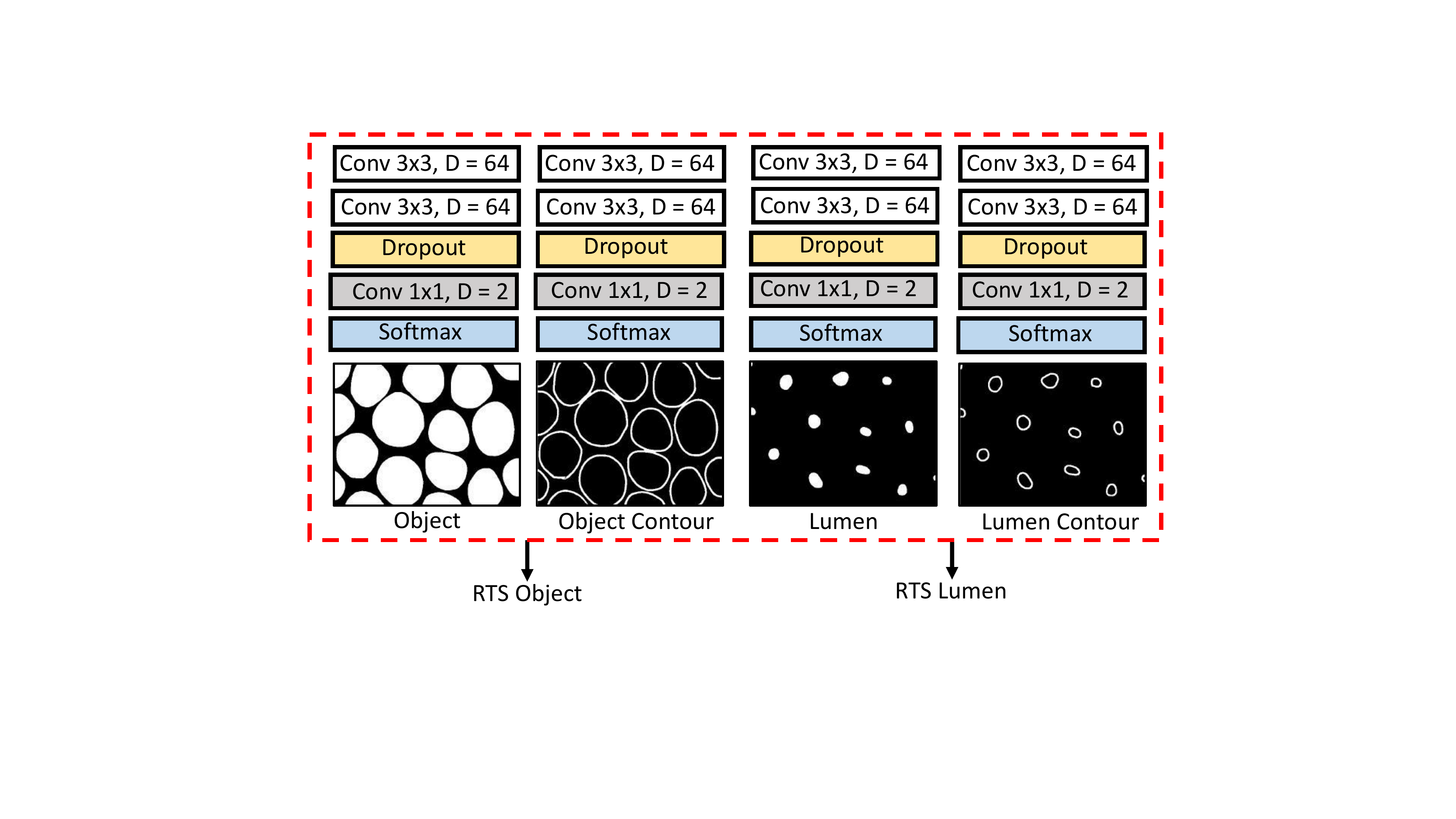} \label{fig:framework}}
\caption*{Figure 3: MILD-Net$^+$. The red dashed box denotes the modified component of MILD-Net$^+$. We observe that the network segments the gland, gland contour, lumen and lumen contour, whilst only applying a small modification to the original network.}
\label{fig:detection}
\end{figure}

\subsection{MILD-Net$^+$ for Simultaneous Gland and Lumen Segmentation}

We extend MILD-Net such that it simultaneously segments the lumen and the gland, in order to increase the diagnostic power of the network. For example, when the grade of cancer increases, tumours tend to become solid and lose their lumenal properties. Therefore, the additional segmentation of the lumen can empower current automated colorectal cancer classification methods, due to the introduction of additional important diagnostic features. In order to achieve this simultanous segmentation, the network requires minimal modification. MILD-Net$^+$ takes an image as input and, identically to MILD-Net, extracts features via the minimal information loss encoder. After upsampling to the original resolution, the task-specific component of the network is modified such that it has four branches. The only difference between MILD-Net and MILD-Net$^+$ is the number of branches after the network is up-sampled back to the size of the original image. Specifically, the part of the architecture shown by the red dashed box displayed in Figure 2(a) is replaced with the one in Figure 3. We observe that the majority of the network is unchanged apart from the addition of two branches at the end of the up-sampling path. As a result, MILD-Net$^+$ does not require many additional parameters to achieve an accurate and simultaneous gland and lumen segmentation. This highlights the ability of MILD-Net$^+$ to extract a rich set of features. Similar to what we have done before, we apply RTS to both the gland and the lumen and use the gland and lumen contours to refine the output of each uncertainty map. Consequently, MILD-Net$^+$ segments diagnostically important features, whilst quantifying the uncertainty for each segmented component.

\subsection{MILD-Net$^+$ Loss Function}

In the same fashion as Section 2.2, we calculate the cross-entropy loss with respect to the output of each component of MILD-Net$^+$. Specifically, we calculate the cross entropy loss with respect to the gland, gland contour, lumen and lumen contour denoted by $\mathcal{L}_g$, $\mathcal{L}_{g_c}$, $\mathcal{L}_l$ and $\mathcal{L}_{l_c}$ respectively. We aso calculate the auxiliary losses $\mathcal{L}_{a_g}$ and $\mathcal{L}_{a_l}$ with respect to the gland and the lumen. Then, during training, the overall loss function of MILD-Net$^+$ is defined as:

\begin{equation}
\mathcal{L}_{total} = \mathcal{L}_g + \mathcal{L}_{gc} +\mathcal{L}_l + \mathcal{L}_{lc} + \lambda\mathcal{L}_{a_g} + \lambda\mathcal{L}_{a_l} +\gamma||\mathbold{\theta}||_2^2
\end{equation}
\noindent where $||\mathbold{\theta}||_2^2$ denotes the regularisation term on weights $\mathbold{\theta}=\{\mathbold{\theta}_g,\mathbold{\theta}_{gc},\mathbold{\theta}_l,\mathbold{\theta}_{lc},\mathbold{\theta}_{a_g},\mathbold{\theta}_{a_l}\}$, with regularisation parameter $\gamma$. We use the same $\gamma$ as MILD-Net, with a value of $10^{-5}$. Also, we use the same $\lambda$ that was utilised within MILD-Net that decays the contribution of the auxiliary loss during training. In a similar vein, we also divide the value by 10 after every eight training epochs. Note, that we choose not include the auxiliary loss with respect to the contours in order to reduce the number of parameters in MILD-Net$^+$.

\section{Experiments and Results}
\subsection{Datasets and Pre-processing}
For our experiments, we used two datasets: (i) the Gland Segmentation (GlaS) challenge dataset~\citep{sirinukunwattana2017gland}, used as part of MICCAI 2015, and (ii) a second independent colon adenocarcinoma dataset, which for simplicity we refer to as the colorectal adenocarcinoma gland (CRAG) dataset\footnote{The CRAG dataset for gland segmentation is available at \url{https://warwick.ac.uk/fac/sci/dcs/research/tia/data/mildnet}}, that was originally used in~\cite{awan2017glandular}. Both datasets were obtained from the University Hospitals Coventry and Warwickshire (UHCW) NHS Trust in Coventry, United Kingdom. Within (i), data was extracted from 16 H\&E stained histological WSIs, scanned with a Zeiss MIRAX MIDI Slide Scanner with a pixel resolution of 0.465$\mu$m/pixel. After scanning, the WSIs were rescaled to 0.620$\mu$m/pixel (equivalent to 20$\times$ objective magnification) and then a total of 165 image tiles were extracted. These 165 images consist of 85 training (37 benign and 48 malignant) and 80 test images (37 benign and 43 malignant). Furthermore, the test images are split into two test sets: Test A and Test B. Test A was released to the participants of the GlaS challenge one month before the submission deadline, whereas Test B was released on the final day of the challenge. Further information on the dataset can be found in the published challenge paper\citep{sirinukunwattana2017gland}. Images are mostly of size 775$\times$522 pixels and all training images have associated instance-level segmentation ground truth that precisely highlight the gland boundaries. In addition, two expert pathologists (D.S, Y.W.T) provide accurate lumen annotations for all glands within the GlaS dataset. Within (ii), we have a total of 213 H\&E CRA images taken from 38 WSIs scanned with an Omnyx VL120 scanner with a pixel resolution of 0.55$\mu$m/pixel (20$\times$ objective magnification). All 38 WSIs are from different patients and are mostly of size 1512$\times$1516 pixels, with corresponding instance-level ground truth. The CRAG dataset is split into 173 training images and 40 test images with different cancer grades. For both datasets, we set 20\% of the training set aside for evaluating the performance of our model during training. Examples of images from each of the two datasets can be seen in Figure 1.

We extracted patches of size 500$\times$500 and augmented patches with elastic distortion, random flip, random rotation, Gaussian blur, median blur and colour distortion. Finally, we randomly cropped a patch of size 464$\times$464, before input into the proposed network.

\subsection{Whole-Slide Image Processing}
In addition to processing the image tiles as described in Section 3.1, we further investigated the ability of our method by processing a set of colorectal adenocarcinoma WSIs. This dataset consists of 16 high resolution WSIs, taken from the COMET dataset, which was originally used in~\cite{sirinukunwattana2016locality}. Within this dataset, the WSIs are obtained from two different centres and therefore we split the images into two further datasets. We name the dataset corresponding to WSIs from the first centre as COMET-1 and the dataset containing WSIs from the second centre as COMET-2. COMET-1 is from the same centre as the image tiles that the algorithm was trained on, whereas COMET-2 is from a different centre completely. We introduce the second centre to test how our method generalises to new data. The data is divided equally, such that 8 WSIs are taken from each centre. Because it is quite laborious to obtain pixel-based glandular annotations for each WSI, we select two high-power fields (HPFs) from each WSI of size 2500$\times$2500 pixels at 20$\times$. As a result, even though we process the whole-slide to see how our algorithm performs visually, we use these selected HPFs to perform quantitative comparison. HPFs were extracted such that we had an even representation of benign and malignant regions, annotated by two expert pathologists (D.S, Y.W.T). In order to satisfy this criteria, we mainly processed WSIs that contained a combination of malignant and benign glands.

\subsection{Implementation Details}
We implemented our framework with the open-source software library TensorFlow version 1.3.0~\citep{abadi2016tensorflow}. We used Xavier initialisation~\citep{glorot2010understanding} for the weights of the model, where they were drawn from a Gaussian distribution. Concretely, weight $w_i$ is initialised with mean 0 and variance $\frac{1}{n_{w_i}}$, where, $n_{w_i}$ refers to the number of input neurons to weight $i$. We trained our model on a workstation equipped with one NVIDIA GEFORCE Titan X GPU for 30 epochs (60,000 steps) on the GlaS dataset and 75 epochs (200,000 steps) on the CRAG dataset. The difference in the number of steps until convergence reflects the greater variability of the CRAG dataset. We used Adam optimisation with an initial learning rate of 10$^{-4}$ and a batch size of 2.

\subsection{Evaluation and Comparison}
We assessed the performance of our method by using the same evaluation criteria used in the MICCAI GlaS challenge, consisting of $F_1$ score, object-level dice and object-level Hausdorff distance~\citep{sirinukunwattana2017gland}. The F1 score is employed to measure the detection accuracy of individual glandular objects, the Dice index is a measure of similarity between two sets of samples and the Hausdorff distance measures the boundary-based segmentation accuracy. We implemented several state-of-the-art segmentation methods including SegNet~\citep{badrinarayanan2015segnet}, FCN-8~\citep{long2015fully} and a DeepLab-v3~\citep{chen2018deeplab} model for extensive comparative analysis. For gland segmentation, we also report the results obtained by two recent methods including MIMO-Net~\citep{raza2017mimonet}, that uses a multi-input-multi-output convolutional neural network and two methods that utilise deep multichannel side supervision~\citep{xu2016gland,xu2017gland}.

For all methods, including MILD-Net, the final binary maps are obtained by applying a threhold of 0.5 to all predicted probability maps. A morphological opening operation is then used with a disk filter radius 5 to obtain the final result. This disk size was emperically selected because it gave the best visual and quantitative results. 

In this section, we first show results for MILD-Net on the GlaS dataset and the CRAG dataset. Next, we display results of MILD-Net for whole-slide image (WSI) processing. Finally, we report results of MILD-Net$^+$ on the GlaS dataset. 

\subsubsection{Results on GlaS and CRAG Datasets Using MILD-Net}
\begin{figure}[!t]
\centering
\captionsetup[subfigure]{labelformat=empty}
\subfloat[]{\includegraphics[width=0.8\columnwidth]{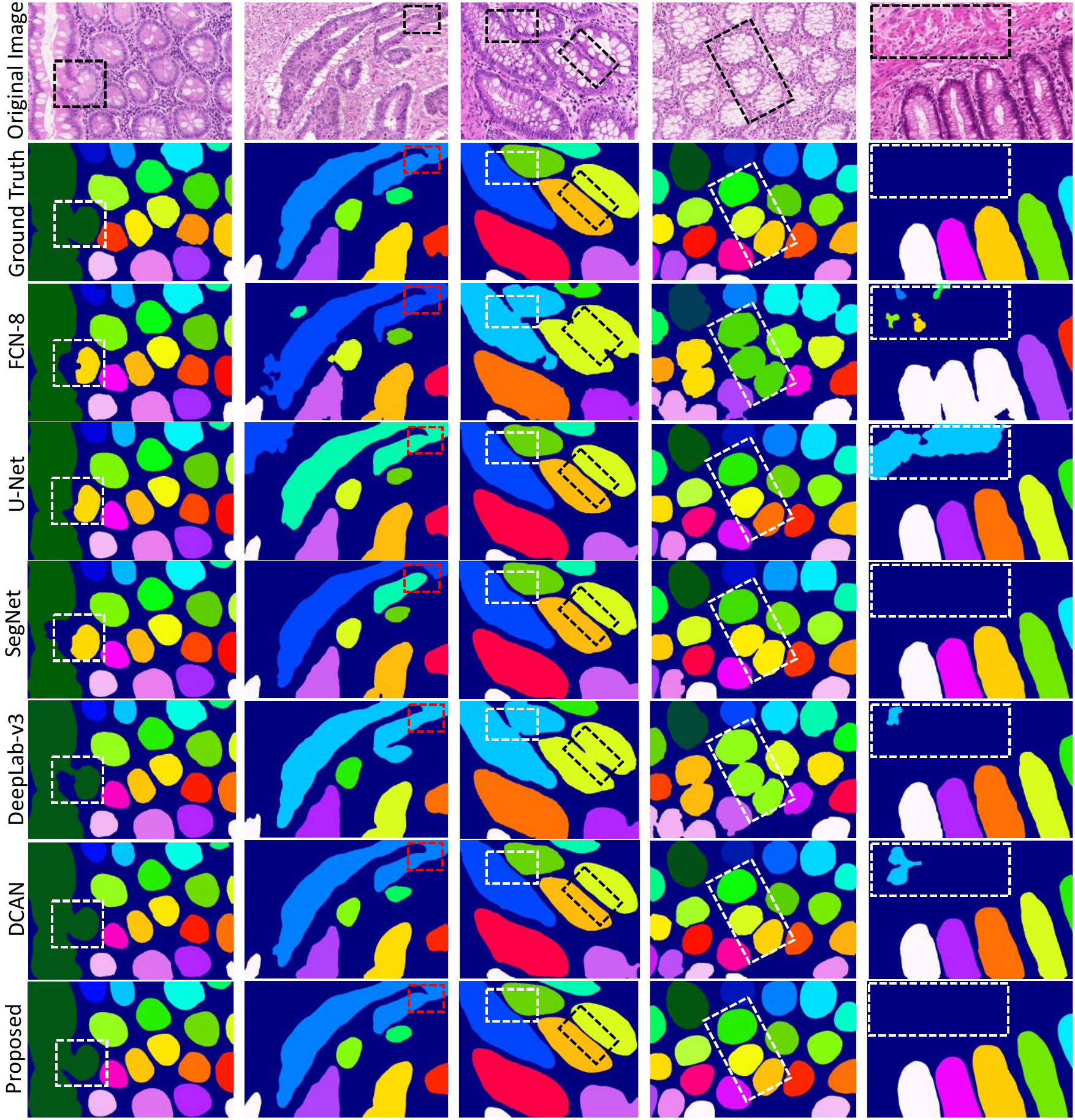} \label{fig:detection1}}
\caption*{Figure 4: Visual gland segmentation results on the GlaS dataset. We compare our method to state-of-the-art methods including FCN-8, U-Net, SegNet, DCAN and DeepLab-v3. Note, visual results for U-Net and DCAN are the results as submitted to the GlaS challenge.}
\end{figure}

\begin{figure}[h!]
\centering
\captionsetup[subfigure]{labelformat=empty}
\subfloat[]{\includegraphics[width=0.8\columnwidth]{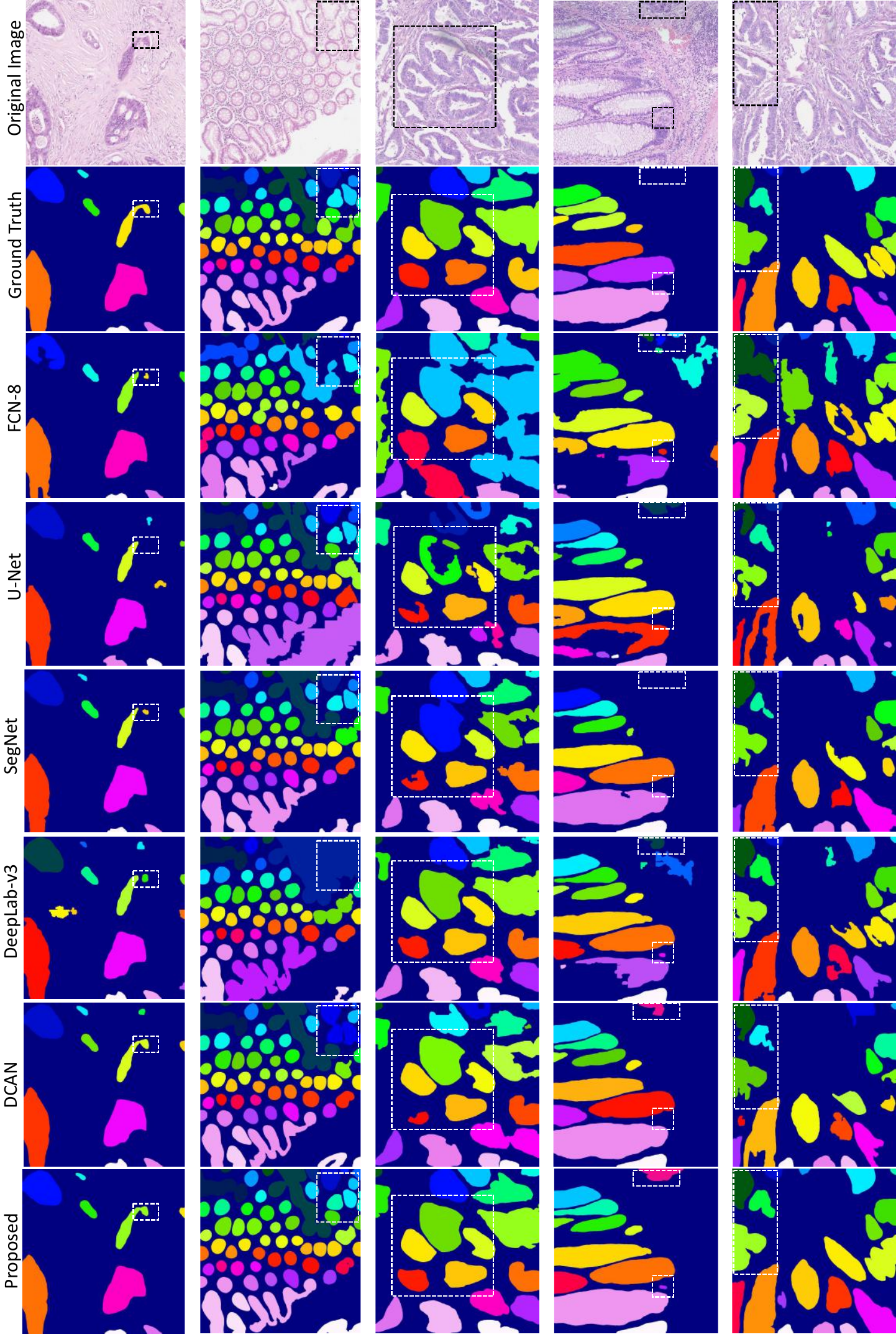} \label{fig:detection1}}
\caption*{Figure 5: Visual gland segmentation results on the CRAG dataset. We compare our method to state-of-the-art methods including FCN-8, U-Net, SegNet, DCAN and DeepLab-v3. }
\end{figure}

\begin{table}[h!]
\small
\label{T:equipos}
\begin{center}
\caption{Comparative analysis of models on the GlaS challenge dataset. CUMedVision submissions use the method reported in \cite{chen2016dcan} and Freidburg submissions use the method reported in \cite{ronneberger2015u}. S and R denote score and rank respectively.}
\begin{tabular}{|p{2.8cm}|c|p{0.25cm}|c|p{0.3cm}|c|p{0.3cm}|c|p{0.3cm}|c|p{0.3cm}|c|p{0.3cm}|p{0.85cm}|}
\hline
\textbf{} & \multicolumn{4}{ c |}{\textbf{$F_1$ Score}} & \multicolumn{4}{ c |}{\textbf{Obj. Dice}} & \multicolumn{4}{ c |}{\textbf{Obj. Hausdorff}} & \\
\cline{2-13}

& \multicolumn{2}{ c |}{\textbf{Test A}} & \multicolumn{2}{ c |}{\textbf{Test B}}& \multicolumn{2}{ c |}{\textbf{Test A}}& \multicolumn{2}{ c |}{\textbf{Test B}}& \multicolumn{2}{ c |}{\textbf{Test A}}& \multicolumn{2}{ c |}{\textbf{Test B}} & \textbf{Rank}  \\
\cline{2-13}

& \textbf{S} & \textbf{R} & \textbf{S} & \textbf{R}& \textbf{S} & \textbf{R} & \textbf{S} & \textbf{R}& \textbf{S} & \textbf{R} & \textbf{S} & \textbf{R}& \textbf{Sum}\\
\hline
 \textbf{MILD-Net} & 0.914 & 1 & 0.844 & 1 & 0.913 & 1 & 0.836 & 1 & 41.54 & 1 & 105.89 & 1 & 6\\
\cite{xu2017gland} & 0.893 & 5 & 0.843 & 2 & 0.908 & 2 & 0.833 & 2 & 44.13 & 2 & 116.82 & 2 & 15\\
MIMO-Net & 0.913 & 2 & 0.724 & 7 & 0.906 & 3 & 0.785 & 9 & 49.15 & 4 & 133.98 & 6 & 31\\
\cite{xu2016gland}  & 0.858 & 11 & 0.771& 3 & 0.888 & 5 & 0.815 & 3 & 54.20 & 5 & 129.93 & 5 & 33\\
DeepLab & 0.862 & 10 & 0.764& 5 & 0.859 & 13 & 0.804 & 5 & 65.72 & 9 & 124.97 & 4& 46\\
SegNet& 0.858 & 11 & 0.753& 6 & 0.864 & 12 & 0.807 & 4 & 62.62 & 10 & 118.51 & 3& 46\\
FCN-8& 0.783 & 14 & 0.692& 12 & 0.795 & 14 & 0.767 & 10 & 105.04 & 12 & 147.28 & 9& 71\\ \hline
CUMedVision2 & 0.912 & 3 & 0.716& 9 & 0.897 & 4 & 0.781 & 11 & 45.42 & 3 & 160.35 & 13 & 43\\
ExB1 & 0.891 & 7 & 0.703& 10 & 0.882 & 8 & 0.786 & 7 & 57.41 & 10 & 145.58 & 7& 49\\
ExB3 & 0.896 & 4 & 0.719& 8 & 0.886 & 6 & 0.765 & 13 & 57.36 & 9 & 159.87 & 12& 52\\
Freidburg2  & 0.87 & 8 & 0.695& 11 & 0.876 & 9 & 0.786 & 7 & 57.09 & 7 & 148.47 & 10 & 52\\
CUMedVision1 & 0.868 & 9 & 0.769& 4 & 0.867 & 11 & 0.8 & 6 & 74.6 & 13 & 153.65 & 11& 54\\
ExB2 & 0.892 & 6 & 0.686& 13 & 0.884 & 7 & 0.754 & 14 & 54.79 & 6 & 187.44 & 15& 61\\
Freidburg1 & 0.834 & 13 & 0.605& 14 & 0.875 & 10 & 0.783 & 10 & 57.19 & 8 & 146.61 & 8 & 63\\
CVML & 0.652 & 16 & 0.541& 15 & 0.644 & 17 & 0.654 & 15 & 155.43 & 17 & 176.24 & 14 & 94\\
LIB & 0.777 & 15 & 0.306& 17 & 0.781 & 15 & 0.617 & 16 & 112.71 & 16 & 190.45 & 16 & 95\\
vision4GlaS & 0.635 & 17 & 0.527& 16 & 0.737 & 16 & 0.61 & 17 & 107.49 & 15 & 210.1 & 17 & 98\\ \hline
\end{tabular}
\end{center}
\end{table}

\begin{table}[t!]
\label{T:equipos}
\small
\begin{center}
\caption{Comparative analysis of models on the CRAG dataset. S and R denote score and rank respectively.}
\begin{tabular}{|p{2.05cm}|c|c|c|c|c|c|c|}
\hline
\textbf{} & \multicolumn{2}{ c |}{\textbf{$F_1$ Score}} & \multicolumn{2}{ c |}{\textbf{Obj. Dice}} & \multicolumn{2}{ c |}{\textbf{Obj. Hausdorff}} & \textbf{Rank}  \\
\cline{2-7}
 & \textbf{S} & \textbf{R} & \textbf{S} & \textbf{R} & \textbf{S} & \textbf{R} & \textbf{Sum}\\\hline

\textbf{MILD-Net} & 0.825 & 1 & 0.875 & 1 & 160.14 & 1 & 3\\
DCAN & 0.736 & 2 & 0.794& 2 & 218.76 & 2 & 6\\
DeepLab & 0.648 & 3 & 0.745& 3 & 281.45 & 4 & 10\\
SegNet & 0.622 & 4 & 0.739& 4 & 247.84 & 3 & 11\\
U-Net & 0.600 & 5 & 0.654& 5 & 354.09 & 5 & 15\\
FCN-8 & 0.558 & 6 & 0.640 & 6 & 436.43 & 6 & 18\\ \hline
\end{tabular}
\end{center}
\end{table}

\begin{table}[t!]
\small
\label{T:equipos}
\begin{center}
\caption{MILD-Net performance with random transformation sampling (RTS) on the CRAG and GlaS datasets.}
\begin{tabular}{|c|c|c|c|c|}
\hline
 \textbf{Dataset} & \textbf{Model}  & \textbf{$F_1$ Score} & \textbf{Obj. Dice} & \textbf{Obj. Hausdorff}\\
\cline{1-5}
GlaS A &  MILD-Net  & 0.914 & 0.908 & 42.32\\
 & MILD-Net-RTS  & 0.914 & \textbf{0.913} & \textbf{41.54} \\ \hline
 GlaS B & MILD-Net &  0.809 & 0.822 & 117.91\\
  & MILD-Net-RTS & \textbf{0.844} & \textbf{0.836} & \textbf{105.89} \\ \hline
CRAG & MILD-Net  & 0.806 & 0.867 & 162.35\\
 & MILD-Net-RTS & \textbf{0.825} & \textbf{0.875} & \textbf{160.14} \\ \hline
\end{tabular}
\end{center}
\end{table}

\begin{figure}[!h]
\centering
\captionsetup[subfigure]{labelformat=empty}
\subfloat[]{\includegraphics[width=0.95\columnwidth]{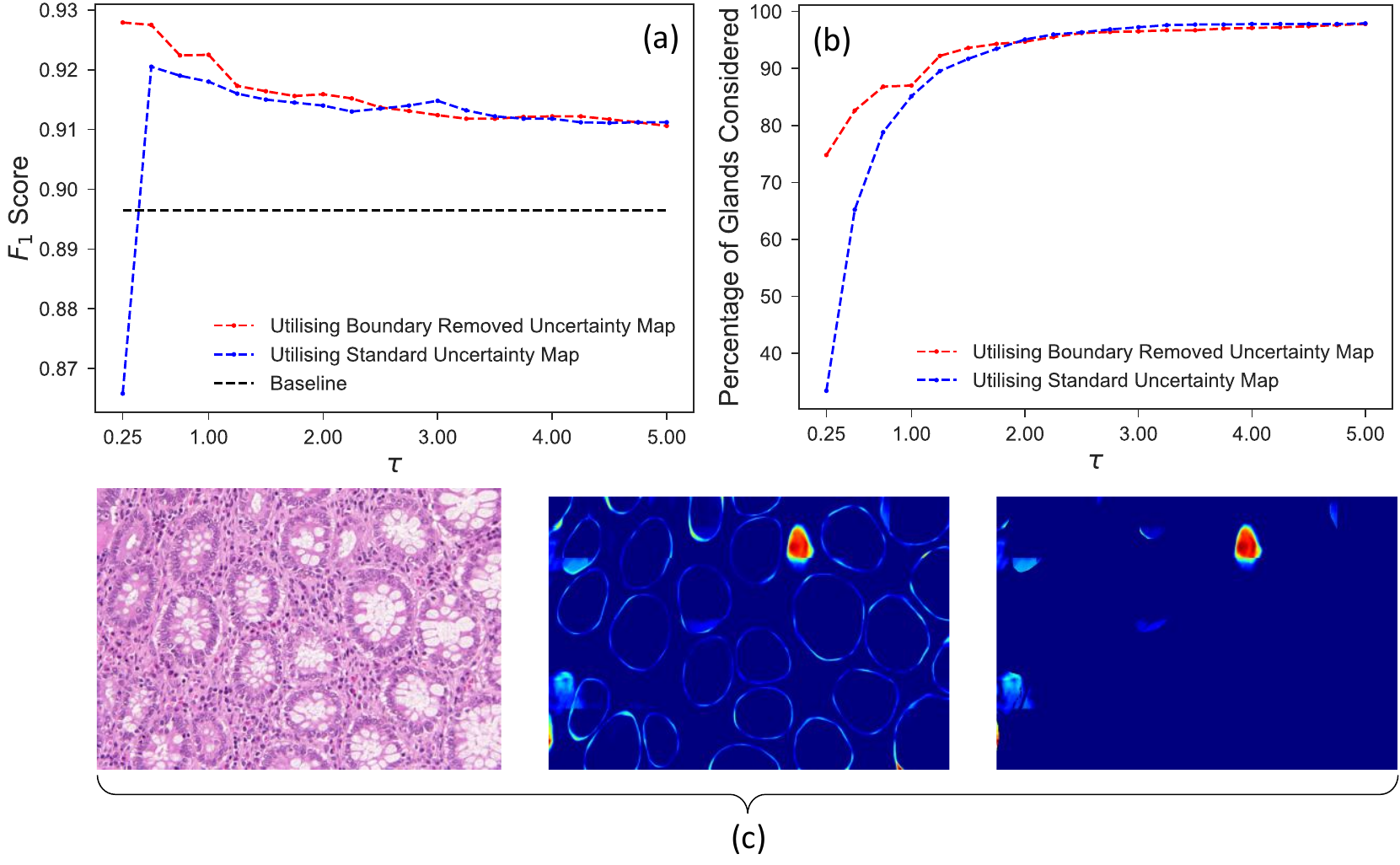} \label{fig:detection1}}
\caption*{Figure 6: Object-level uncertainty quantification. (a) shows the $F_1$ score as we disregard predictions with an uncertainty score $\tau_k$ greater than a given threshold $\tau$. (b) The percentage of total instances considered, given a threshold $\tau$. For the red dashed line, we use the boundary removed uncertainty map, whereas for the blue dashed line we use the standard uncertainty map. The black horizontal line shows the $F_1$ score when no glands with a high uncertainty are removed. (a) and (b) relate to results on the combined set of test A and test B. (c) from left to right: original image; uncertainty map $\sigma$; boundary removed uncertainty map $\hat{\sigma}$. For each instance $k$ within $\hat{\sigma}$, an object-level uncertainty score $\tau$ is calculated. }
\end{figure}

We can see from Table 1 that our proposed network achieves state-of-the-art performance compared to all methods on the 2015 MICCAI GlaS Challenge dataset.
We also validated the efficacy of our method on the CRAG dataset, demonstrating overall better performance in comparison with other methods and highlighting the good generalisation capability of our method on different datasets. Results on the CRAG dataset can be seen in Table 2. We can see from Table 3 that utilising test time random transformations leads to an improved performance, due to a refined prediction within areas of high uncertainty. Additionally, we compared our method of RTS to Monte Carlo dropout sampling. However, because we don't apply many dropout layers within our network, there is not sufficient variation in the samples to have a profound effect. We also experimented by adding additional dropout layers with Monte Carlo dropout, but this had a detrimental effect during the training of the network. Because RTS utilises an averaging technique, the number of false positives in areas of high uncertainty is reduced. This explains the increase in performance with RTS. It must be noted that it is significantly more difficult to segment glands within the CRAG dataset than when using the GlaS dataset. This is because there are many malignant cases where the glandular boundaries are very ambiguous. Examples of results from different methods are shown in Figure 4 and 5. We can see that our method can generate more accurate gland instance segmentation with precisely delineated boundaries and well segmented instances. It is interesting to see that within the dashed boxes in the last column of Figure 4, our proposed algorithm was able to detect tumorous areas that were not picked up by the pathologist. 

In Figure 6, we show the relationship between the performance and the uncertainty score $\tau$. This score is used as a threshold, where we only consider predictions $k$ with an uncertainty score $\tau_k$ lower than $\tau$. We observe from Figure 6 that it seems sensible to only consider segmented predictions with an uncertainty score $\tau_k$ below 1. This preserves a large proportion of the dataset, whilst significantly increasing the performance. We also display the effect of using the boundary removed uncertainty map. We observe that removing the boundary allows us to preserve a larger proportion of the dataset when we are using lower thresholds for the removal of predictions with high uncertainty. This suggests that using the boundary removed uncertainty map allows us to correctly remove the uncertain cases that contribute most negatively to the performance. Therefore, utilising the boundary removed uncertainty map is more robust and can be effectively be used to select predictions with low uncertainty. It is interesting to note that we are still able to preserve around 75\% of instances by selecting predictions with $\tau_k$ below 0.25. As a result, $F_1$ score, object dice and object Hausdorff can be increased to 0.930, 0.9359 and 28.658 for test set A and increased to 0.913, 0.9567 and 22.70 for test set B. It must be noted that the intuition of disregarding glands with high uncertainty means that we should not extract any statistical measures from these disregarded glands. Therefore, when removing predicted instances with high uncertainty, we also remove the corresponding ground truth instance to obtain the above measures. 

\subsubsection{Results on Whole-Slide Images Using MILD-Net}

Within part (a-d) of Figure 7, the inner-most image is the original WSI with overlaid glandular boundaries, the central column shows the two HPFs for statistical analysis at 20$\times$ and the outer-most column shows a selected region of each HPF at 40$\times$. We observe that our proposed method is able to accurately segment glands within colon whole-slide histology images with a precise delineation of glandular boundaries. Therefore, as a result of training on both the GlaS and the CRAG dataset, our method is capable of extracting a strong set of features that enables a successful transition to WSI processing. We also note from part (c) and (d) of Figure 7, that MILD-Net generalises well to completely unseen data from different centres. A particularly interesting aspect of COMET-2 is that most images contain pathologist pen markings. However, as a result of the strong set of features that MILD-Net is able to extract no pre-processing was needed to avoid these regions, where other methods may have failed. For a thorough analysis, we obtain quantitative results for all HPFs extracted from the 16 WSIs. In total, we have 32 HPFs: 16 from COMET-1 and 16 from COMET-2. In order to test the performance of our algorithm on both benign and malignant cases, we ensured an equal representation of both benign and malignant glands. We can see from Table 4 that the proposed method has a similar performance between the two datasets, highlighting the generalisability of our method. Despite a good detection performance, we can see that the Hausdorff distance within malignant cases is significantly higher than those results reported on the GlaS and the CRAG dataset. The Hausdorff distance measure indicates how closely the shape of two objects match with each other. As a result, disagreement at the boundary will lead to deterioration in performance. Therefore, this suggests that the algorithm finds it challenging to precisely locate the glandular boundaries within malignant cases. This however reflects the true difficulty in segmenting glands within whole-slide histology images, where there are often many ambiguous regions. After careful observation, we state that the lower performance for Hausdorff distance is not due to a limitation of the algorithm, but because a number of mailgnant cases are generally difficult to segment. 
\begin{figure}[h!]
\centering
\captionsetup[subfigure]{labelformat=empty}
\subfloat[]{\includegraphics[width=0.99\columnwidth]{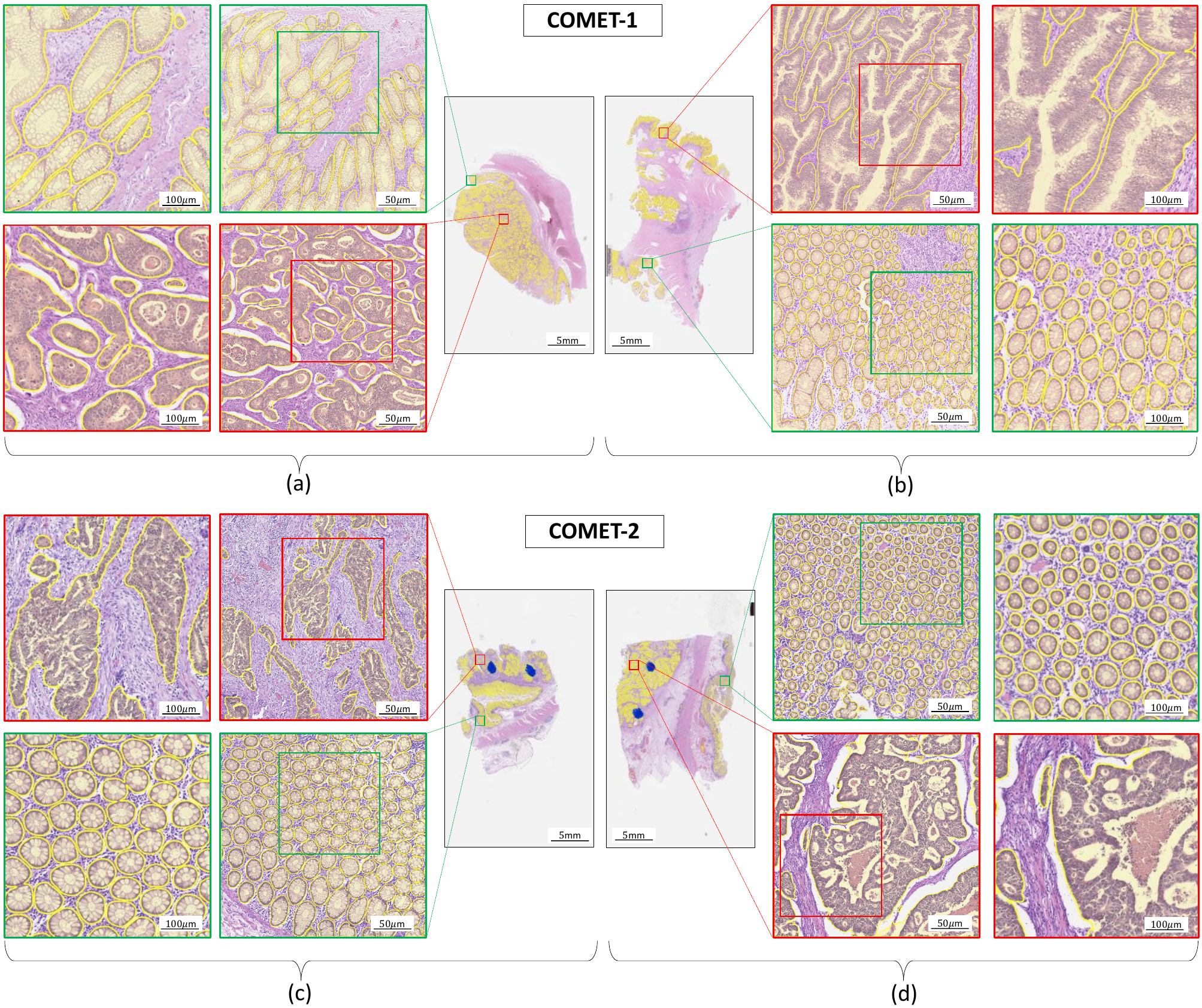} \label{fig:detection1}}
\caption*{Figure 7: Visual results of gland segmentation in WSIs using the proposed framework. (a) and (b) show processed images using COMET-1, whilst (c) and (d) show processed images using COMET-2. Red regions show malignant areas of interest, whereas green regions show benign areas of interest. The central column of images within (a), (b), (c) and (d) shows the two HPFs extracted from each WSI for statistical analysis.}
\end{figure}

\begin{table}[h!]
\small
\label{T:equipos}
\begin{center}
\caption{MILD-Net gland segmentation performance on HPFs from WSIs. B stands for average benign score and M stands for average malignant score.}
\begin{tabular}{|c|c|c|c|c|c|c|}
\hline
& \multicolumn{2}{ | c |}{\textbf{$F_1$ Score}} & \multicolumn{2}{ c |}{\textbf{Obj. Dice}} & \multicolumn{2}{ c |}{\textbf{Obj. Hausdorff}}\\
\cline{2-7}
& \textbf{B} & \textbf{M} & \textbf{B} & \textbf{M} & \textbf{B} & \textbf{M}\\
\hline
\textbf{COMET-1} & 0.811 & 0.817 &  0.822 & 0.867 & 158.40  & 389.89 \\ \hline
\textbf{COMET-2} & 0.948 & 0.716 & 0.886 & 0.751 & 76.15  & 474.12 \\ \hline
\textbf{Average COMET-1} & \multicolumn{2}{ c |}{0.814}& \multicolumn{2}{ c |}{0.845} & \multicolumn{2}{ c |}{274.15} \\ \hline
\textbf{Average COMET-2} & \multicolumn{2}{ c |}{0.832}& \multicolumn{2}{ c |}{0.819} & \multicolumn{2}{ c |}{275.14} \\ \hline
\end{tabular}
\end{center}
\end{table}

\subsubsection{Results on GlaS and CRAG Datasets Using MILD-Net$^+$}
To demonstrate the performance of MILD-Net$^+$, we compare our algorithm to four recent segmentation methods trained solely for the task of lumen segmentation. Namely, these methods are FCN-8~\citep{long2015fully}, U-Net~\citep{ronneberger2015u}, SegNet~\citep{badrinarayanan2015segnet} and DeepLab-v3~\citep{chen2018deeplab}. We chose not to compare with DCAN~\citep{chen2016dcan} because this network was specifically tuned to achieve instance segmentation. Instance segmentation is not an issue for lumen segmentation, because neighbouring lumen physically can't touch within histology images. The only exception for this would be if there was an artifact within the image. From Figure 8, we observe that our algorithm is able to precisely segment both the gland object and the gland lumen. We can see in Table 5, that MILD-Net$^+$ achieves superior performance in all statistical measures for lumen segmentation, compared to all competing methods. This is particularly interesting because all other competing methods were trained for the single task of lumen segmentation. Therefore, this reiterates the strong feature extraction capabilities of the minimal information loss network. Despite achieving state-of-the-art performance at the output of the lumen branch, it is necessary to ensure that we still achieve a good accuracy at the output of the gland object branch. We observe that, MILD-Net$^+$ out-performs MILD-Net on most of the statistical measures, suggesting that segmenting the lumen may provide additional cues to strengthen the segmentation of the gland object.

\begin{figure}[!t]
\centering
\captionsetup[subfigure]{labelformat=empty}
\subfloat[]{\includegraphics[width=0.99\columnwidth]{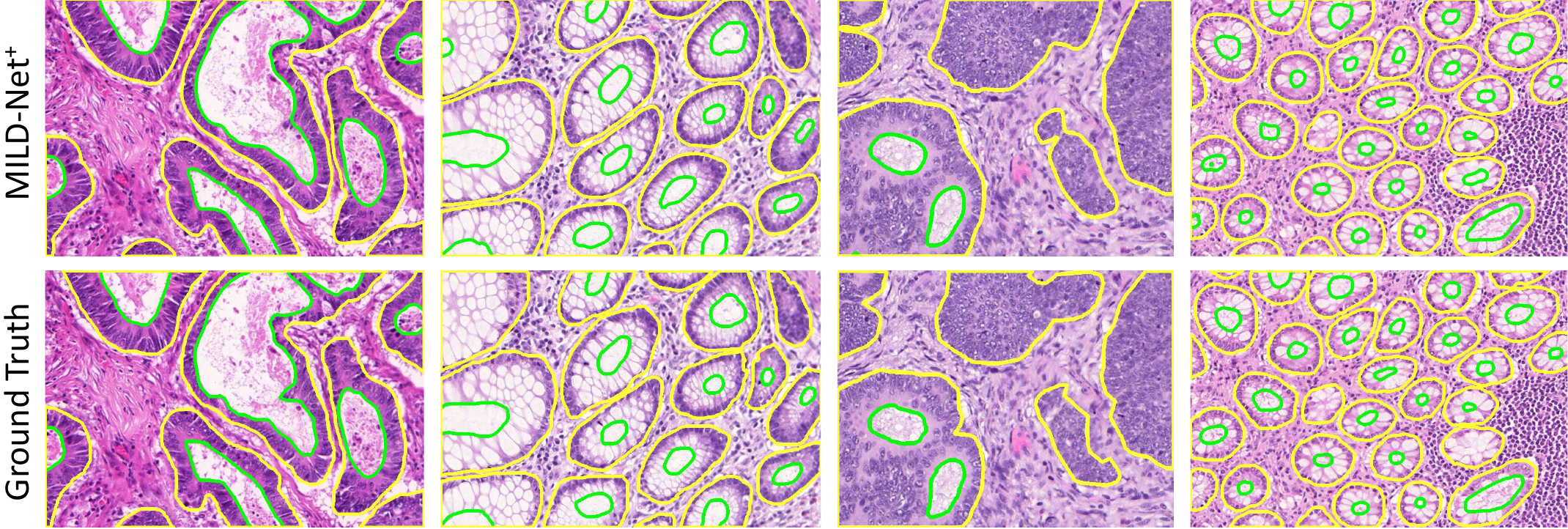} \label{fig:detection1}}
\caption*{Figure 8: Visual results of gland and lumen segmentation. The top row displays the output of the proposed method. The bottom row displays the pathologist annotation. Yellow contours show the outline the glandular boundaries and green contours show the outline of the lumen boundaries.}
\end{figure}

\begin{table}[h!]
\small
\label{T:equipos}
\begin{center}
\caption{MILD-Net$^+$ lumen segmentation performance on the GlaS challenge dataset}
\begin{tabular}{|p{0.1cm}|c|p{1.06cm}|p{1.06cm}|p{1.06cm}|p{1.06cm}|p{1.06cm}|p{1.06cm}|}
\hline
\multicolumn{2}{ |c }{\textbf{}}& \multicolumn{2}{ | c |}{\textbf{$F_1$ Score}} & \multicolumn{2}{ c |}{\textbf{Obj. Dice}} & \multicolumn{2}{ c |}{\textbf{Obj. Hausdorff}}\\
\cline{3-8}
\multicolumn{2}{ |c| }{\textbf{}} & \textbf{Test A} & \textbf{Test B} & \textbf{Test A} & \textbf{Test B} & \textbf{Test A} & \textbf{Test B}\\
\hline
 \parbox[t]{2mm}{\multirow{5}{*}{\rotatebox[origin=c]{90}{\textbf{Lumen}}}}& \textbf{MILD-Net$^+$} & 0.825  & 0.711  & 0.875 & 0.816 & 26.81 & 94.09\\
& DeepLab & 0.757  & 0.521 & 0.816 & 0.722 & 46.49 & 136.81\\
& SegNet & 0.698 & 0.661 & 0.791 & 0.781 & 56.22 & 110.32\\
& U-Net  & 0.623 & 0.425 & 0.724 & 0.643 & 73.51 & 152.52\\
& FCN-8 & 0.744 & 0.556 & 0.778 & 0.723 & 60.51  & 133.09 \\ \hline
\parbox[t]{2mm}{\multirow{3}{*}{\rotatebox[origin=c]{90}{\textbf{Gland}}}}
&  \textbf{MILD-Net$^+$} & 0.920  & 0.820  & 0.918 & 0.836 & 39.39 & 103.07 \\
& \textbf{MILD-Net} & 0.914  & 0.844  & 0.913 & 0.836 & 41.54 & 105.89 \\
& CUMedVision2& 0.912 & 0.716 & 0.897 & 0.781 & 45.42 & 160.35 \\
 \hline
\end{tabular}
\end{center}
\end{table}

\section{Discussion}

Analysis of Hematoxylin and Eosin stained histology slides is considered as the "gold standard" in histology based diagnosis. However, visual analysis is very time consuming and laborious because pathologists are required to thoroughly examine each case to ensure an accurate diagnosis. Furthermore, due to the complex nature of the task, histopathological diagnosis often suffers from inter- and intra-observer variability. Computational techniques aim to counter the challenges posed within routine pathology by providing an objective and potentially more accurate diagnosis. In order to improve the diagnostic capabilities of automated methods, we present a minimal information loss dilated network for the accurate segmentation of glands within colon histology images. Subsequently, gland based features can be used to empower the diagnostic decision made by the pathologist. 

Extensive experimentation on multiple datasets demonstrates the superior performance of our approach compared to other competing methods. Furthermore, our method performs well when applied to the WSI, highlighting the network's strong feature extraction capabilities. As a result, the network may be used in a clinical setting to segment glandular structures within the WSI with a high level of accuracy. We also show that the method generalises well to new data and can therefore be expected to work well within other centres. 

It is worth noting, that the minimal information loss network helps retain the spatial information within the network and therefore leads to a successful segmentation at the glandular boundaries. Therefore, additional cues are not needed to separate the majority of touching instances. However, it must be noted that this method is able to separate glands when they are very close together, but may fail when the glands are physically touching with no pixels in between. We do not see this as a cause for concern because the majority of instances can be separated by our method due to the reduction of information loss throughout the network. We also observe from our results that our network was able to successfully segment glands of various sizes. This in part was because of the addition of the \textit{atrous} spatial pyramid pooling module that enlarged the size of the receptive field with varying dilation rates.

The addition of RTS increased the performance of the algorithm, whilst simultaneously generating an uncertainty map. We have shown that this uncertainty map can be used as additional information about where the algorithm is uncertain. Also, we have shown that if we choose not to extract features from predictions with high uncertainty, we can signifcantly increase the performance whilst maintaining a large proportion of the dataset. We can ensure that we retain a larger proportion of this dataset if we use a boundary removed uncertainty map. The removal of predictions with high uncertainty is particularly important for gland-based feature extraction (e.g glandular aberrance~\citep{awan2017glandular}) because features should not be extracted from glands where the algorithm is not confident. This workflow mimics clinical practice because the pathologist would not make a diagnosis from areas of ambiguity. Therefore, this uncertainty map can be used to extract relatively strong features for subsequent grading. 

The proposed network may fail to distinguish between the lumen of the gastrointestinal tract and the glandular lumen. However, this is to be expected because of a very similar appearance between these histological components. As well as this, we only used small image tiles for developing our algorithm and therefore contextual information to empower the segmentation is limited. In future work, we may incorporate a larger input size to provide additional context to the algorithm. 

With a small modification, the network is able to precisely segment the lumen of the gland. We also observed that the segmentation is very accurate within benign glands. This is positive because we presumed that there may have been confusion between lumenal areas and areas containing goblet cells. After performing this segmentation, lumenal features can be used to empower current automated classification methods, that are limited to features extracted from solely the gland object. We also observe that the additional segmentation of the lumen leads to an overall superior gland segmentation. This suggests that the lumen can provide additional cues to help increase the overall performance of gland instance segmentation.

In future work we will develop our proposed method for successful and fast whole-slide image processing. Therefore, we aim to adapt our method such that it can process a WSI in a short amount of time, whilst maintaining a similar level of accuracy. Our current method utilises RTS for uncertainty map generation. Although this uncertainty map is very informative, we must develop a non-ensembling approach if we plan to efficiently process the WSI in a short amount of time. As well as this, we will develop an effective pre-processing pipeline to ensure non-informative regions are not processed. On another note, it must be made clear that this algorithm is currently limited to colon cancer because of the data that it was trained on. The work could be extended such that we are able to segment the glands within other tissue, given that we have sufficient data.

\section{Conclusion}
In this paper, we presented a minimal information loss dilated network for gland instance segmentation in colon histology images. The proposed network retains maximal information during feature extraction that is very important for successful gland instance segmentation. Furthermore, in order to segment glands of various sizes, we use \textit{atrous} spatial pyramid pooling for effective multi-scale aggregation. To incorporate uncertainty within our framework, we apply random transformations to images during test time. Taking the average of this sample leads to a superior segmentation, whilst simultaneously allowing us to visualise areas of ambiguity. Furthermore, we propose an object-level uncertainty score that can be used for assessing whether to discard predictions with high uncertainty. We also highlight the generalisability of our method by processing whole-slide images from a different centre with high accuracy. As an extension, we show how our proposed method can be adapted such that it simultaneously segments the gland lumen and the gland object. 
We observe that our method obtains state-of-the-art performance in the MICCAI 2015 gland segmentation challenge and on a second independent colorectal adenocarcinoma dataset.  

\vspace{-0.5em}

\section{Acknowledgements}
An earlier version of this work was presented at the Medical Imaging with Deep Learning (MIDL) conference in Amsterdam in July 2018. The authors are grateful to the Warwick Global Partnership Fund (GPF) 2017/18 for funding this collaboration between Warwick and CUHK. H.C, Q.D and P.-A.H are supported by the Hong Kong Innovation and Technology Commission, under ITSP Tier 3 (project number: ITS/041/16).

\vspace{-0.5em}


\bibliography{mild_net}

\begin{thebibliography}{44}
\expandafter\ifx\csname natexlab\endcsname\relax\def\natexlab#1{#1}\fi
\providecommand{\url}[1]{\texttt{#1}}
\providecommand{\href}[2]{#2}
\providecommand{\path}[1]{#1}
\providecommand{\DOIprefix}{doi:}
\providecommand{\ArXivprefix}{arXiv:}
\providecommand{\URLprefix}{URL: }
\providecommand{\Pubmedprefix}{pmid:}
\providecommand{\doi}[1]{\href{http://dx.doi.org/#1}{\path{#1}}}
\providecommand{\Pubmed}[1]{\href{pmid:#1}{\path{#1}}}
\providecommand{\bibinfo}[2]{#2}
\ifx\xfnm\relax \def\xfnm[#1]{\unskip,\space#1}\fi
\bibitem[{Abadi et~al.(2016)Abadi, Barham, Chen, Chen, Davis, Dean, Devin,
  Ghemawat, Irving, Isard et~al.}]{abadi2016tensorflow}
\bibinfo{author}{Abadi, M.}, \bibinfo{author}{Barham, P.},
  \bibinfo{author}{Chen, J.}, \bibinfo{author}{Chen, Z.},
  \bibinfo{author}{Davis, A.}, \bibinfo{author}{Dean, J.},
  \bibinfo{author}{Devin, M.}, \bibinfo{author}{Ghemawat, S.},
  \bibinfo{author}{Irving, G.}, \bibinfo{author}{Isard, M.}, et~al.,
  \bibinfo{year}{2016}.
\newblock \bibinfo{title}{Tensorflow: a system for large-scale machine
  learning.}, in: \bibinfo{booktitle}{OSDI}, pp. \bibinfo{pages}{265--283}.
\bibitem[{Albarqouni et~al.(2016)Albarqouni, Baur, Achilles, Belagiannis,
  Demirci and Navab}]{albarqouni2016aggnet}
\bibinfo{author}{Albarqouni, S.}, \bibinfo{author}{Baur, C.},
  \bibinfo{author}{Achilles, F.}, \bibinfo{author}{Belagiannis, V.},
  \bibinfo{author}{Demirci, S.}, \bibinfo{author}{Navab, N.},
  \bibinfo{year}{2016}.
\newblock \bibinfo{title}{Aggnet: deep learning from crowds for mitosis
  detection in breast cancer histology images}.
\newblock \bibinfo{journal}{IEEE transactions on medical imaging}
  \bibinfo{volume}{35}, \bibinfo{pages}{1313--1321}.
\bibitem[{Awan et~al.(2017)Awan, Sirinukunwattana, Epstein, Jefferyes, Qidwai,
  Aftab, Mujeeb, Snead and Rajpoot}]{awan2017glandular}
\bibinfo{author}{Awan, R.}, \bibinfo{author}{Sirinukunwattana, K.},
  \bibinfo{author}{Epstein, D.}, \bibinfo{author}{Jefferyes, S.},
  \bibinfo{author}{Qidwai, U.}, \bibinfo{author}{Aftab, Z.},
  \bibinfo{author}{Mujeeb, I.}, \bibinfo{author}{Snead, D.},
  \bibinfo{author}{Rajpoot, N.}, \bibinfo{year}{2017}.
\newblock \bibinfo{title}{Glandular morphometrics for objective grading of
  colorectal adenocarcinoma histology images}.
\newblock \bibinfo{journal}{Scientific reports} \bibinfo{volume}{7},
  \bibinfo{pages}{16852}.
\bibitem[{Badrinarayanan et~al.(2015)Badrinarayanan, Kendall and
  Cipolla}]{badrinarayanan2015segnet}
\bibinfo{author}{Badrinarayanan, V.}, \bibinfo{author}{Kendall, A.},
  \bibinfo{author}{Cipolla, R.}, \bibinfo{year}{2015}.
\newblock \bibinfo{title}{Segnet: A deep convolutional encoder-decoder
  architecture for image segmentation}.
\newblock \bibinfo{journal}{arXiv preprint arXiv:1511.00561} .
\bibitem[{Bejnordi et~al.(2017)Bejnordi, Veta, Van~Diest, Van~Ginneken,
  Karssemeijer, Litjens, Van Der~Laak, Hermsen, Manson, Balkenhol
  et~al.}]{bejnordi2017diagnostic}
\bibinfo{author}{Bejnordi, B.E.}, \bibinfo{author}{Veta, M.},
  \bibinfo{author}{Van~Diest, P.J.}, \bibinfo{author}{Van~Ginneken, B.},
  \bibinfo{author}{Karssemeijer, N.}, \bibinfo{author}{Litjens, G.},
  \bibinfo{author}{Van Der~Laak, J.A.}, \bibinfo{author}{Hermsen, M.},
  \bibinfo{author}{Manson, Q.F.}, \bibinfo{author}{Balkenhol, M.}, et~al.,
  \bibinfo{year}{2017}.
\newblock \bibinfo{title}{Diagnostic assessment of deep learning algorithms for
  detection of lymph node metastases in women with breast cancer}.
\newblock \bibinfo{journal}{Jama} \bibinfo{volume}{318},
  \bibinfo{pages}{2199--2210}.
\bibitem[{Bishop(2006)}]{bishop2012pattern}
\bibinfo{author}{Bishop, C.M.}, \bibinfo{year}{2006}.
\newblock \bibinfo{title}{Pattern recognition and machine learning. springer} .
\bibitem[{Chen et~al.(2016a)Chen, Dou, Wang, Qin, Heng
  et~al.}]{chen2016mitosis}
\bibinfo{author}{Chen, H.}, \bibinfo{author}{Dou, Q.}, \bibinfo{author}{Wang,
  X.}, \bibinfo{author}{Qin, J.}, \bibinfo{author}{Heng, P.A.}, et~al.,
  \bibinfo{year}{2016}a.
\newblock \bibinfo{title}{Mitosis detection in breast cancer histology images
  via deep cascaded networks.}, in: \bibinfo{booktitle}{AAAI}, pp.
  \bibinfo{pages}{1160--1166}.
\bibitem[{Chen et~al.(2017)Chen, Qi, Yu, Dou, Qin and Heng}]{chen2017dcan}
\bibinfo{author}{Chen, H.}, \bibinfo{author}{Qi, X.}, \bibinfo{author}{Yu, L.},
  \bibinfo{author}{Dou, Q.}, \bibinfo{author}{Qin, J.}, \bibinfo{author}{Heng,
  P.A.}, \bibinfo{year}{2017}.
\newblock \bibinfo{title}{Dcan: Deep contour-aware networks for object instance
  segmentation from histology images}.
\newblock \bibinfo{journal}{Medical image analysis} \bibinfo{volume}{36},
  \bibinfo{pages}{135--146}.
\bibitem[{Chen et~al.(2016b)Chen, Qi, Yu and Heng}]{chen2016dcan}
\bibinfo{author}{Chen, H.}, \bibinfo{author}{Qi, X.}, \bibinfo{author}{Yu, L.},
  \bibinfo{author}{Heng, P.A.}, \bibinfo{year}{2016}b.
\newblock \bibinfo{title}{Dcan: deep contour-aware networks for accurate gland
  segmentation}, in: \bibinfo{booktitle}{Proceedings of the IEEE conference on
  Computer Vision and Pattern Recognition}, pp. \bibinfo{pages}{2487--2496}.
\bibitem[{Chen et~al.(2018)Chen, Papandreou, Kokkinos, Murphy and
  Yuille}]{chen2018deeplab}
\bibinfo{author}{Chen, L.C.}, \bibinfo{author}{Papandreou, G.},
  \bibinfo{author}{Kokkinos, I.}, \bibinfo{author}{Murphy, K.},
  \bibinfo{author}{Yuille, A.L.}, \bibinfo{year}{2018}.
\newblock \bibinfo{title}{Deeplab: Semantic image segmentation with deep
  convolutional nets, atrous convolution, and fully connected crfs}.
\newblock \bibinfo{journal}{IEEE transactions on pattern analysis and machine
  intelligence} \bibinfo{volume}{40}, \bibinfo{pages}{834--848}.
\bibitem[{Cire{\c{s}}an et~al.(2013)Cire{\c{s}}an, Giusti, Gambardella and
  Schmidhuber}]{cirecsan2013mitosis}
\bibinfo{author}{Cire{\c{s}}an, D.C.}, \bibinfo{author}{Giusti, A.},
  \bibinfo{author}{Gambardella, L.M.}, \bibinfo{author}{Schmidhuber, J.},
  \bibinfo{year}{2013}.
\newblock \bibinfo{title}{Mitosis detection in breast cancer histology images
  with deep neural networks}, in: \bibinfo{booktitle}{International Conference
  on Medical Image Computing and Computer-assisted Intervention},
  \bibinfo{organization}{Springer}. pp. \bibinfo{pages}{411--418}.
\bibitem[{Compton(2000)}]{compton2000updated}
\bibinfo{author}{Compton, C.C.}, \bibinfo{year}{2000}.
\newblock \bibinfo{title}{Updated protocol for the examination of specimens
  from patients with carcinomas of the colon and rectum, excluding carcinoid
  tumors, lymphomas, sarcomas, and tumors of the vermiform appendix: a basis
  for checklists}.
\newblock \bibinfo{journal}{Archives of pathology \& laboratory medicine}
  \bibinfo{volume}{124}, \bibinfo{pages}{1016--1025}.
\bibitem[{Fleming et~al.(2012)Fleming, Ravula, Tatishchev and
  Wang}]{fleming2012colorectal}
\bibinfo{author}{Fleming, M.}, \bibinfo{author}{Ravula, S.},
  \bibinfo{author}{Tatishchev, S.F.}, \bibinfo{author}{Wang, H.L.},
  \bibinfo{year}{2012}.
\newblock \bibinfo{title}{Colorectal carcinoma: pathologic aspects}.
\newblock \bibinfo{journal}{Journal of gastrointestinal oncology}
  \bibinfo{volume}{3}, \bibinfo{pages}{153}.
\bibitem[{Gal(2016)}]{Gal2016Uncertainty}
\bibinfo{author}{Gal, Y.}, \bibinfo{year}{2016}.
\newblock \bibinfo{title}{Uncertainty in Deep Learning}.
\newblock Ph.D. thesis. University of Cambridge.
\bibitem[{Gal and Ghahramani(2016)}]{gal2016dropout}
\bibinfo{author}{Gal, Y.}, \bibinfo{author}{Ghahramani, Z.},
  \bibinfo{year}{2016}.
\newblock \bibinfo{title}{Dropout as a bayesian approximation: Representing
  model uncertainty in deep learning}, in: \bibinfo{booktitle}{international
  conference on machine learning}, pp. \bibinfo{pages}{1050--1059}.
\bibitem[{Glorot and Bengio(2010)}]{glorot2010understanding}
\bibinfo{author}{Glorot, X.}, \bibinfo{author}{Bengio, Y.},
  \bibinfo{year}{2010}.
\newblock \bibinfo{title}{Understanding the difficulty of training deep
  feedforward neural networks}, in: \bibinfo{booktitle}{Proceedings of the
  thirteenth international conference on artificial intelligence and
  statistics}, pp. \bibinfo{pages}{249--256}.
\bibitem[{Graham and Rajpoot(2018)}]{graham2018sams}
\bibinfo{author}{Graham, S.}, \bibinfo{author}{Rajpoot, N.M.},
  \bibinfo{year}{2018}.
\newblock \bibinfo{title}{Sams-net: Stain-aware multi-scale network for
  instance-based nuclei segmentation in histology images}, in:
  \bibinfo{booktitle}{2018 IEEE 15th International Symposium on Biomedical
  Imaging (ISBI 2018)}, \bibinfo{organization}{IEEE}. pp.
  \bibinfo{pages}{590--594}.
\bibitem[{Graham et~al.(2018)Graham, Shaban, Qaiser, Khurram and
  Rajpoot}]{graham2018classification}
\bibinfo{author}{Graham, S.}, \bibinfo{author}{Shaban, M.},
  \bibinfo{author}{Qaiser, T.}, \bibinfo{author}{Khurram, S.A.},
  \bibinfo{author}{Rajpoot, N.}, \bibinfo{year}{2018}.
\newblock \bibinfo{title}{Classification of lung cancer histology images using
  patch-level summary statistics}, in: \bibinfo{booktitle}{Medical Imaging
  2018: Digital Pathology}, \bibinfo{organization}{International Society for
  Optics and Photonics}. p. \bibinfo{pages}{1058119}.
\bibitem[{Gurcan et~al.(2009)Gurcan, Boucheron, Can, Madabhushi, Rajpoot and
  Yener}]{gurcan2009histopathological}
\bibinfo{author}{Gurcan, M.N.}, \bibinfo{author}{Boucheron, L.},
  \bibinfo{author}{Can, A.}, \bibinfo{author}{Madabhushi, A.},
  \bibinfo{author}{Rajpoot, N.}, \bibinfo{author}{Yener, B.},
  \bibinfo{year}{2009}.
\newblock \bibinfo{title}{Histopathological image analysis: A review}.
\newblock \bibinfo{journal}{IEEE reviews in biomedical engineering}
  \bibinfo{volume}{2}, \bibinfo{pages}{147}.
\bibitem[{Hamilton et~al.(2000)Hamilton, Aaltonen
  et~al.}]{hamilton2000pathology}
\bibinfo{author}{Hamilton, S.R.}, \bibinfo{author}{Aaltonen, L.A.}, et~al.,
  \bibinfo{year}{2000}.
\newblock \bibinfo{title}{Pathology and genetics of tumours of the digestive
  system}. volume~\bibinfo{volume}{48}.
\newblock \bibinfo{publisher}{IARC press Lyon:}.
\bibitem[{Kendall and Gal(2017)}]{kendall2017uncertainties}
\bibinfo{author}{Kendall, A.}, \bibinfo{author}{Gal, Y.}, \bibinfo{year}{2017}.
\newblock \bibinfo{title}{What uncertainties do we need in bayesian deep
  learning for computer vision?}, in: \bibinfo{booktitle}{Advances in neural
  information processing systems}, pp. \bibinfo{pages}{5574--5584}.
\bibitem[{Kong et~al.(2017)Kong, Wang, Li, Song and Zhang}]{kong2017cancer}
\bibinfo{author}{Kong, B.}, \bibinfo{author}{Wang, X.}, \bibinfo{author}{Li,
  Z.}, \bibinfo{author}{Song, Q.}, \bibinfo{author}{Zhang, S.},
  \bibinfo{year}{2017}.
\newblock \bibinfo{title}{Cancer metastasis detection via spatially structured
  deep network}, in: \bibinfo{booktitle}{International Conference on
  Information Processing in Medical Imaging}, \bibinfo{organization}{Springer}.
  pp. \bibinfo{pages}{236--248}.
\bibitem[{LeCun et~al.(2015)LeCun, Bengio and Hinton}]{lecun2015deep}
\bibinfo{author}{LeCun, Y.}, \bibinfo{author}{Bengio, Y.},
  \bibinfo{author}{Hinton, G.}, \bibinfo{year}{2015}.
\newblock \bibinfo{title}{Deep learning}.
\newblock \bibinfo{journal}{nature} \bibinfo{volume}{521},
  \bibinfo{pages}{436}.
\bibitem[{Lin et~al.(2018)Lin, Chen, Dou, Wang, Qin and Heng}]{lin2018scannet}
\bibinfo{author}{Lin, H.}, \bibinfo{author}{Chen, H.}, \bibinfo{author}{Dou,
  Q.}, \bibinfo{author}{Wang, L.}, \bibinfo{author}{Qin, J.},
  \bibinfo{author}{Heng, P.A.}, \bibinfo{year}{2018}.
\newblock \bibinfo{title}{Scannet: A fast and dense scanning framework for
  metastastic breast cancer detection from whole-slide image}, in:
  \bibinfo{booktitle}{2018 IEEE Winter Conference on Applications of Computer
  Vision (WACV)}, \bibinfo{organization}{IEEE}. pp. \bibinfo{pages}{539--546}.
\bibitem[{Litjens et~al.(2017)Litjens, Kooi, Bejnordi, Setio, Ciompi,
  Ghafoorian, Van Der~Laak, Van~Ginneken and S{\'a}nchez}]{litjens2017survey}
\bibinfo{author}{Litjens, G.}, \bibinfo{author}{Kooi, T.},
  \bibinfo{author}{Bejnordi, B.E.}, \bibinfo{author}{Setio, A.A.A.},
  \bibinfo{author}{Ciompi, F.}, \bibinfo{author}{Ghafoorian, M.},
  \bibinfo{author}{Van Der~Laak, J.A.}, \bibinfo{author}{Van~Ginneken, B.},
  \bibinfo{author}{S{\'a}nchez, C.I.}, \bibinfo{year}{2017}.
\newblock \bibinfo{title}{A survey on deep learning in medical image analysis}.
\newblock \bibinfo{journal}{Medical image analysis} \bibinfo{volume}{42},
  \bibinfo{pages}{60--88}.
\bibitem[{Long et~al.(2015)Long, Shelhamer and Darrell}]{long2015fully}
\bibinfo{author}{Long, J.}, \bibinfo{author}{Shelhamer, E.},
  \bibinfo{author}{Darrell, T.}, \bibinfo{year}{2015}.
\newblock \bibinfo{title}{Fully convolutional networks for semantic
  segmentation}, in: \bibinfo{booktitle}{Proceedings of the IEEE conference on
  computer vision and pattern recognition}, pp. \bibinfo{pages}{3431--3440}.
\bibitem[{Nalisnick and Smyth(2018)}]{nalisnick2018learning}
\bibinfo{author}{Nalisnick, E.}, \bibinfo{author}{Smyth, P.},
  \bibinfo{year}{2018}.
\newblock \bibinfo{title}{Learning priors for invariance}, in:
  \bibinfo{booktitle}{International Conference on Artificial Intelligence and
  Statistics}, pp. \bibinfo{pages}{366--375}.
\bibitem[{Qaiser et~al.(2018)Qaiser, Mukherjee, Reddy~Pb, Munugoti, Tallam,
  Pitk{\"a}aho, Lehtim{\"a}ki, Naughton, Berseth, Pedraza
  et~al.}]{qaiser2018her}
\bibinfo{author}{Qaiser, T.}, \bibinfo{author}{Mukherjee, A.},
  \bibinfo{author}{Reddy~Pb, C.}, \bibinfo{author}{Munugoti, S.D.},
  \bibinfo{author}{Tallam, V.}, \bibinfo{author}{Pitk{\"a}aho, T.},
  \bibinfo{author}{Lehtim{\"a}ki, T.}, \bibinfo{author}{Naughton, T.},
  \bibinfo{author}{Berseth, M.}, \bibinfo{author}{Pedraza, A.}, et~al.,
  \bibinfo{year}{2018}.
\newblock \bibinfo{title}{Her 2 challenge contest: a detailed assessment of
  automated her 2 scoring algorithms in whole slide images of breast cancer
  tissues}.
\newblock \bibinfo{journal}{Histopathology} \bibinfo{volume}{72},
  \bibinfo{pages}{227--238}.
\bibitem[{Qaiser et~al.(2017)Qaiser, Tsang, Epstein and
  Rajpoot}]{qaiser2017tumor}
\bibinfo{author}{Qaiser, T.}, \bibinfo{author}{Tsang, Y.W.},
  \bibinfo{author}{Epstein, D.}, \bibinfo{author}{Rajpoot, N.},
  \bibinfo{year}{2017}.
\newblock \bibinfo{title}{Tumor segmentation in whole slide images using
  persistent homology and deep convolutional features}, in:
  \bibinfo{booktitle}{Annual Conference on Medical Image Understanding and
  Analysis}, \bibinfo{organization}{Springer}. pp. \bibinfo{pages}{320--329}.
\bibitem[{Raza et~al.(2017)Raza, Cheung, Epstein, Pelengaris, Khan and
  Rajpoot}]{raza2017mimonet}
\bibinfo{author}{Raza, S.E.A.}, \bibinfo{author}{Cheung, L.},
  \bibinfo{author}{Epstein, D.}, \bibinfo{author}{Pelengaris, S.},
  \bibinfo{author}{Khan, M.}, \bibinfo{author}{Rajpoot, N.M.},
  \bibinfo{year}{2017}.
\newblock \bibinfo{title}{Mimonet: Gland segmentation using
  multi-input-multi-output convolutional neural network}, in:
  \bibinfo{booktitle}{Annual Conference on Medical Image Understanding and
  Analysis}, \bibinfo{organization}{Springer}. pp. \bibinfo{pages}{698--706}.
\bibitem[{Ronneberger et~al.(2015)Ronneberger, Fischer and
  Brox}]{ronneberger2015u}
\bibinfo{author}{Ronneberger, O.}, \bibinfo{author}{Fischer, P.},
  \bibinfo{author}{Brox, T.}, \bibinfo{year}{2015}.
\newblock \bibinfo{title}{U-net: Convolutional networks for biomedical image
  segmentation}, in: \bibinfo{booktitle}{International Conference on Medical
  image computing and computer-assisted intervention},
  \bibinfo{organization}{Springer}. pp. \bibinfo{pages}{234--241}.
\bibitem[{Sabour et~al.(2017)Sabour, Frosst and Hinton}]{sabour2017dynamic}
\bibinfo{author}{Sabour, S.}, \bibinfo{author}{Frosst, N.},
  \bibinfo{author}{Hinton, G.E.}, \bibinfo{year}{2017}.
\newblock \bibinfo{title}{Dynamic routing between capsules}, in:
  \bibinfo{booktitle}{Advances in Neural Information Processing Systems}, pp.
  \bibinfo{pages}{3856--3866}.
\bibitem[{Sapkota et~al.(2018)Sapkota, Shi, Xing and Yang}]{sapkota2018deep}
\bibinfo{author}{Sapkota, M.}, \bibinfo{author}{Shi, X.},
  \bibinfo{author}{Xing, F.}, \bibinfo{author}{Yang, L.}, \bibinfo{year}{2018}.
\newblock \bibinfo{title}{Deep convolutional hashing for low dimensional binary
  embedding of histopathological images}.
\newblock \bibinfo{journal}{IEEE Journal of Biomedical and Health Informatics}
  .
\bibitem[{Shen et~al.(2017)Shen, Wu and Suk}]{shen2017deep}
\bibinfo{author}{Shen, D.}, \bibinfo{author}{Wu, G.}, \bibinfo{author}{Suk,
  H.I.}, \bibinfo{year}{2017}.
\newblock \bibinfo{title}{Deep learning in medical image analysis}.
\newblock \bibinfo{journal}{Annual review of biomedical engineering}
  \bibinfo{volume}{19}, \bibinfo{pages}{221--248}.
\bibitem[{Shi et~al.(2017)Shi, Xing, Xu, Xie, Su and Yang}]{shi2017supervised}
\bibinfo{author}{Shi, X.}, \bibinfo{author}{Xing, F.}, \bibinfo{author}{Xu,
  K.}, \bibinfo{author}{Xie, Y.}, \bibinfo{author}{Su, H.},
  \bibinfo{author}{Yang, L.}, \bibinfo{year}{2017}.
\newblock \bibinfo{title}{Supervised graph hashing for histopathology image
  retrieval and classification}.
\newblock \bibinfo{journal}{Medical image analysis} \bibinfo{volume}{42},
  \bibinfo{pages}{117--128}.
\bibitem[{Sirinukunwattana et~al.(2017)Sirinukunwattana, Pluim, Chen, Qi, Heng,
  Guo, Wang, Matuszewski, Bruni, Sanchez et~al.}]{sirinukunwattana2017gland}
\bibinfo{author}{Sirinukunwattana, K.}, \bibinfo{author}{Pluim, J.P.},
  \bibinfo{author}{Chen, H.}, \bibinfo{author}{Qi, X.}, \bibinfo{author}{Heng,
  P.A.}, \bibinfo{author}{Guo, Y.B.}, \bibinfo{author}{Wang, L.Y.},
  \bibinfo{author}{Matuszewski, B.J.}, \bibinfo{author}{Bruni, E.},
  \bibinfo{author}{Sanchez, U.}, et~al., \bibinfo{year}{2017}.
\newblock \bibinfo{title}{Gland segmentation in colon histology images: The
  glas challenge contest}.
\newblock \bibinfo{journal}{Medical image analysis} \bibinfo{volume}{35},
  \bibinfo{pages}{489--502}.
\bibitem[{Sirinukunwattana et~al.(2016)Sirinukunwattana, Raza, Tsang, Snead,
  Cree and Rajpoot}]{sirinukunwattana2016locality}
\bibinfo{author}{Sirinukunwattana, K.}, \bibinfo{author}{Raza, S.E.A.},
  \bibinfo{author}{Tsang, Y.W.}, \bibinfo{author}{Snead, D.R.},
  \bibinfo{author}{Cree, I.A.}, \bibinfo{author}{Rajpoot, N.M.},
  \bibinfo{year}{2016}.
\newblock \bibinfo{title}{Locality sensitive deep learning for detection and
  classification of nuclei in routine colon cancer histology images}.
\newblock \bibinfo{journal}{IEEE transactions on medical imaging}
  \bibinfo{volume}{35}, \bibinfo{pages}{1196--1206}.
\bibitem[{Veta et~al.(2015)Veta, Van~Diest, Willems, Wang, Madabhushi,
  Cruz-Roa, Gonzalez, Larsen, Vestergaard, Dahl et~al.}]{veta2015assessment}
\bibinfo{author}{Veta, M.}, \bibinfo{author}{Van~Diest, P.J.},
  \bibinfo{author}{Willems, S.M.}, \bibinfo{author}{Wang, H.},
  \bibinfo{author}{Madabhushi, A.}, \bibinfo{author}{Cruz-Roa, A.},
  \bibinfo{author}{Gonzalez, F.}, \bibinfo{author}{Larsen, A.B.},
  \bibinfo{author}{Vestergaard, J.S.}, \bibinfo{author}{Dahl, A.B.}, et~al.,
  \bibinfo{year}{2015}.
\newblock \bibinfo{title}{Assessment of algorithms for mitosis detection in
  breast cancer histopathology images}.
\newblock \bibinfo{journal}{Medical image analysis} \bibinfo{volume}{20},
  \bibinfo{pages}{237--248}.
\bibitem[{Washington et~al.(2009)Washington, Berlin, Branton, Burgart, Carter,
  Fitzgibbons, Halling, Frankel, Jessup, Kakar et~al.}]{washington2009protocol}
\bibinfo{author}{Washington, M.K.}, \bibinfo{author}{Berlin, J.},
  \bibinfo{author}{Branton, P.}, \bibinfo{author}{Burgart, L.J.},
  \bibinfo{author}{Carter, D.K.}, \bibinfo{author}{Fitzgibbons, P.L.},
  \bibinfo{author}{Halling, K.}, \bibinfo{author}{Frankel, W.},
  \bibinfo{author}{Jessup, J.}, \bibinfo{author}{Kakar, S.}, et~al.,
  \bibinfo{year}{2009}.
\newblock \bibinfo{title}{Protocol for the examination of specimens from
  patients with primary carcinoma of the colon and rectum}.
\newblock \bibinfo{journal}{Archives of pathology \& laboratory medicine}
  \bibinfo{volume}{133}, \bibinfo{pages}{1539--1551}.
\bibitem[{Xu et~al.(2016)Xu, Li, Liu, Wang, Lai, Eric and Chang}]{xu2016gland}
\bibinfo{author}{Xu, Y.}, \bibinfo{author}{Li, Y.}, \bibinfo{author}{Liu, M.},
  \bibinfo{author}{Wang, Y.}, \bibinfo{author}{Lai, M.}, \bibinfo{author}{Eric,
  I.}, \bibinfo{author}{Chang, C.}, \bibinfo{year}{2016}.
\newblock \bibinfo{title}{Gland instance segmentation by deep multichannel side
  supervision}, in: \bibinfo{booktitle}{International Conference on Medical
  Image Computing and Computer-Assisted Intervention},
  \bibinfo{organization}{Springer}. pp. \bibinfo{pages}{496--504}.
\bibitem[{Xu et~al.(2017)Xu, Li, Wang, Liu, Fan, Lai, Eric and
  Chang}]{xu2017gland}
\bibinfo{author}{Xu, Y.}, \bibinfo{author}{Li, Y.}, \bibinfo{author}{Wang, Y.},
  \bibinfo{author}{Liu, M.}, \bibinfo{author}{Fan, Y.}, \bibinfo{author}{Lai,
  M.}, \bibinfo{author}{Eric, I.}, \bibinfo{author}{Chang, C.},
  \bibinfo{year}{2017}.
\newblock \bibinfo{title}{Gland instance segmentation using deep multichannel
  neural networks}.
\newblock \bibinfo{journal}{IEEE Transactions on Biomedical Engineering}
  \bibinfo{volume}{64}, \bibinfo{pages}{2901--2912}.
\bibitem[{Yang et~al.(2017)Yang, Zhang, Chen, Zhang and
  Chen}]{yang2017suggestive}
\bibinfo{author}{Yang, L.}, \bibinfo{author}{Zhang, Y.}, \bibinfo{author}{Chen,
  J.}, \bibinfo{author}{Zhang, S.}, \bibinfo{author}{Chen, D.Z.},
  \bibinfo{year}{2017}.
\newblock \bibinfo{title}{Suggestive annotation: A deep active learning
  framework for biomedical image segmentation}, in:
  \bibinfo{booktitle}{International Conference on Medical Image Computing and
  Computer-Assisted Intervention}, \bibinfo{organization}{Springer}. pp.
  \bibinfo{pages}{399--407}.
\bibitem[{Yu and Koltun(2015)}]{yu2015multi}
\bibinfo{author}{Yu, F.}, \bibinfo{author}{Koltun, V.}, \bibinfo{year}{2015}.
\newblock \bibinfo{title}{Multi-scale context aggregation by dilated
  convolutions}.
\newblock \bibinfo{journal}{arXiv preprint arXiv:1511.07122} .
\bibitem[{Zhang et~al.(2017)Zhang, Yang, Chen, Fredericksen, Hughes and
  Chen}]{zhang2017deep}
\bibinfo{author}{Zhang, Y.}, \bibinfo{author}{Yang, L.}, \bibinfo{author}{Chen,
  J.}, \bibinfo{author}{Fredericksen, M.}, \bibinfo{author}{Hughes, D.P.},
  \bibinfo{author}{Chen, D.Z.}, \bibinfo{year}{2017}.
\newblock \bibinfo{title}{Deep adversarial networks for biomedical image
  segmentation utilizing unannotated images}, in:
  \bibinfo{booktitle}{International Conference on Medical Image Computing and
  Computer-Assisted Intervention}, \bibinfo{organization}{Springer}. pp.
  \bibinfo{pages}{408--416}.

\end{thebibliography}

\end{document}